\documentclass[11pt]{article}

\usepackage[preprint]{acl}

\usepackage{times}
\usepackage{latexsym}

\usepackage[T1]{fontenc}
\usepackage[utf8]{inputenc}

\usepackage{microtype}

\usepackage{inconsolata}

\usepackage{graphicx}
\usepackage{microtype}      % microtypography
\usepackage{xcolor}         % colors
\usepackage{enumitem}
\usepackage{multirow}
\usepackage{mathtools}
\usepackage{amssymb}
\usepackage{colortbl}
\usepackage{graphicx}
\usepackage{url}
\usepackage{pgfplots}  
\usepackage{caption}      
\usepackage{subcaption} 
\usepackage{algorithm}
\usepackage{algpseudocode}
\usepackage{amsmath}
\usepackage{booktabs}
\usepackage{tabularx}
\usepackage{wrapfig}
\usepackage{float}
\usepackage{enumitem}
\makeatletter
\renewcommand*{\@fnsymbol}[1]{\ensuremath{\ifcase#1\or \dagger\or \ddagger
\else\@ctrerr\fi}}
\makeatother

\usepackage{tikz}

\newcommand{\ours}{{GRASP}}
\definecolor{RowHighlight}{gray}{0.92}

\title{GRASP: Reinforcing Language Model Anonymizers with Group Relative Policy Optimization}

  \author{Sajjad Ghiasvand, Mark Beliaev, Mahnoosh Alizadeh, \& Ramtin Pedarsani \\
Department of Electrical and Computer Engineering\\
 UC Santa Barbara\\
Santa Barbara, CA 93106, USA \\
\texttt{\{{sajjad,mbeliaev,alizadeh,ramtin\}@ucsb.edu}
}
}

\author{
        \textbf{Sajjad Ghiasvand}$^1$ \ \
        \textbf{Nader Sehatbakhsh}$^2$ \ \
       \\
  Electrical and Computer Engineering Department, UC Santa Barbara$^1$ \ \ \\ Electrical and Computer Engineering Department, UC Los Angeles$^2$
  \\
  {\tt sajjad@ucsb.edu}\ \
  {\tt nsehat@ee.ucla.edu}
}

\begin{document}
\maketitle

\begin{abstract}
Large language models can infer sensitive personal attributes, such as age, location, and occupation, from ordinary text, turning everyday writing into a privacy risk. Adversarial anonymization defends against this by rewriting a text with a capable language model that also plays the attacker, but it needs a powerful model at inference time and thus sends private text to a third party, the very exposure anonymization should prevent. Recent work distills this behavior into a small on-device model using supervised fine-tuning and direct preference optimization (DPO), but DPO only imitates the teacher's offline choices and never directly optimizes the privacy--utility objective we care about. We introduce \textbf{\ours{}} (\textbf{G}roup-\textbf{R}elative \textbf{A}nonymization via \textbf{S}elf-refinement \textbf{P}olicy-optimization), which reinforces the local anonymizer online with Group Relative Policy Optimization. A single small model acts as anonymizer, adversary, and utility judge, trained against a self-generated reward that hides attributes while preserving meaning, with a design that guards against reward hacking. Trained on Llama-3.1-8B, \ours{} improves the privacy--utility
trade-off over the DPO-distilled baseline, consistently across three
independent LLM judges. Against adversarial anonymization driven by
frontier models such as Gemini~2.5~Flash and Claude, it achieves a
comparable or better overall trade-off while removing substantially
more private information, and it runs entirely on-device at roughly
$1\%$ of the GPT-4o teacher's cost.

\end{abstract}

\section{Introduction}
\label{s:intro}
\begin{figure*}[t]
    \centering
    \includegraphics[width=0.9\textwidth]{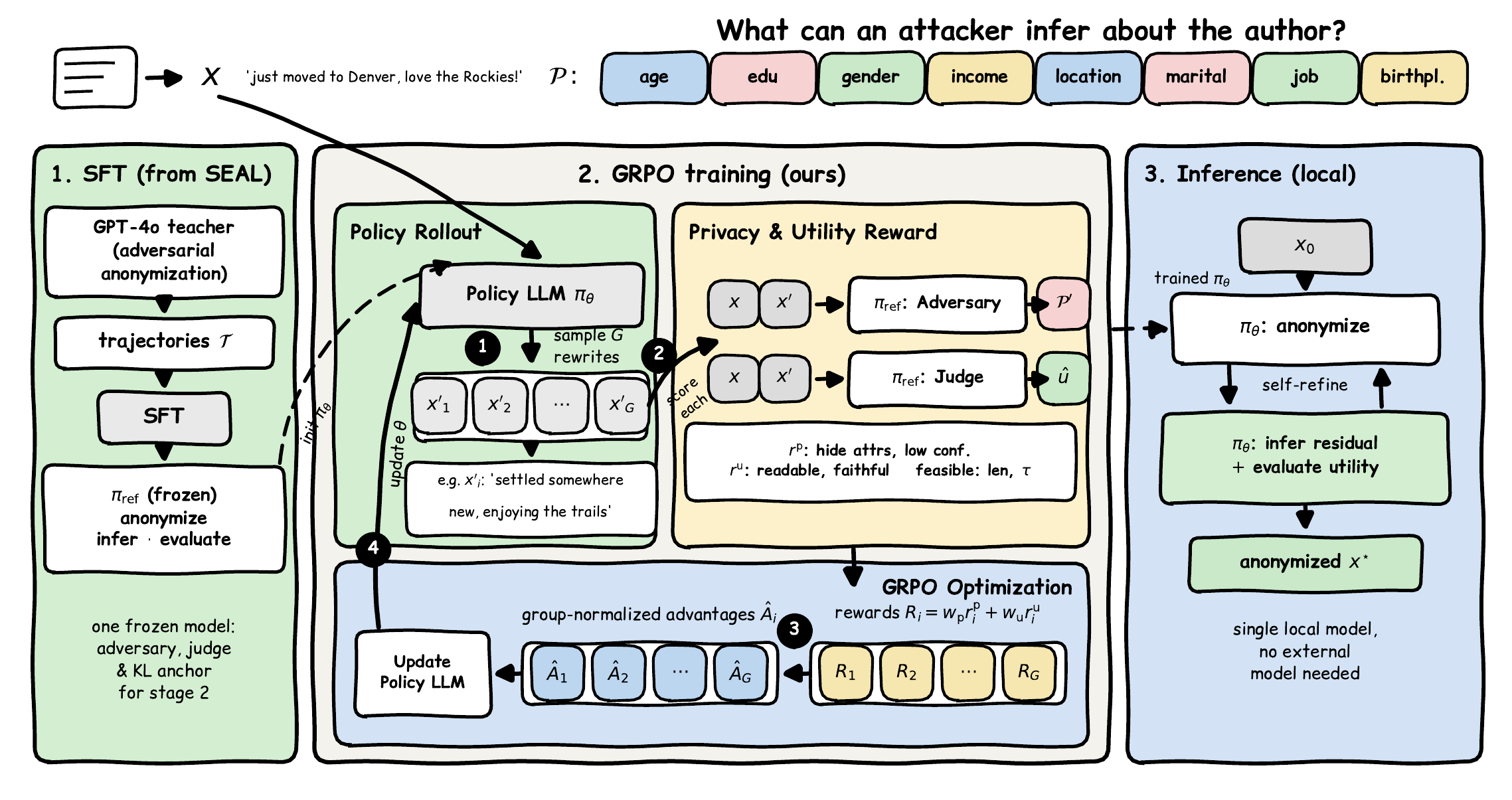}
    \caption{\textbf{Overview of \ours{}.} \emph{Stage~1:} we reuse SEAL's GPT-4o adversarial-anonymization trajectories to supervise-fine-tune a small language model into a frozen reference $\pi_{\text{ref}}$ that can anonymize, infer attributes, and evaluate utility. \emph{Stage~2:} we optimize the anonymization policy $\pi_\theta$ with GRPO. For each text, $\pi_\theta$ samples a group of candidate rewrites; the frozen $\pi_{\text{ref}}$ scores each as adversary and judge into a privacy--utility reward, and group-relative advantages update the policy with a KL anchor to $\pi_{\text{ref}}$. \emph{Stage~3:} at inference the trained policy runs entirely on-device, self-refining by alternately anonymizing and critiquing its own output, with no external model.}
    \label{fig:workflow}
\end{figure*}

Large language models can infer sensitive personal attributes, such as age, location, occupation, and income, from ordinary text with surprising accuracy~\citep{staab2023beyond, staab2024language}, turning routine online writing into a privacy liability and compounding known exposures such as memorization and extraction of training data~\citep{carlini2021extracting, carlini2022quantifying}. As people increasingly write with and share text through LLMs~\citep{achiam2023gpt, grattafiori2024llama, team2025gemma}, there is a growing need to rewrite text so that its meaning is preserved while private attributes can no longer be inferred~\citep{dou2023reducing}.

Classical anonymization removes explicit identifiers through named-entity recognition or rule-based redaction~\citep{microsoft2025, pilan2022tab}, and paraphrasing models alter surface form~\citep{krishna2023paraphrasing}, but neither addresses the contextual cues that modern inference attacks exploit~\citep{gadotti2024anonymization}. The strongest defense to date is adversarial anonymization~\citep{staab2024language, yang2024robust, frikha2024incognitext}, which iteratively rewrites a text using a capable LLM that also plays the attacker. This is effective but requires a powerful model at inference time, so users must either pay for a frontier API or send private text to a third party~\citep{hou2025general}, which is precisely the exposure that anonymization is meant to prevent.

To keep anonymization local, SEAL~\citep{kim2026self} distills~\citep{hinton2015distilling,levine2025mi} adversarial anonymization into a small language model that can anonymize, infer, and evaluate its own rewrites, using supervised fine-tuning followed by direct preference optimization (DPO)~\citep{rafailov2023direct}. DPO, however, only imitates preferences mined offline from the teacher's trajectories: it learns to reproduce which rewrites the teacher preferred, but never directly optimizes the privacy--utility objective we care about, and it cannot improve beyond the demonstrations it distilled from.

Our key observation is that anonymization comes with a well-defined and measurable objective, namely whether an adversary can still infer the target attributes and whether the text's utility is preserved, and that a single distilled model can act as anonymizer, adversary, and utility judge. This makes it natural to optimize the objective \emph{directly} with online reinforcement learning rather than imitate a teacher offline. Whereas reinforcement learning from human feedback trains language models against costly human preference labels~\citep{christiano2017deep, ziegler2019fine, stiennon2020learning, ouyang2022training}, our reward is generated by the model itself, as it iteratively refines its own rewrites~\citep{madaan2023self} and scores them as an attacker and an LLM judge~\citep{zheng2023judging}.

We introduce \textbf{\ours{}} (\textbf{G}roup-\textbf{R}elative \textbf{A}nonymization via \textbf{S}elf-refinement \textbf{P}olicy-optimization), which fine-tunes the anonymization policy with Group Relative Policy Optimization (GRPO)~\citep{shao2024deepseekmath, guo2025deepseek} against a reward that combines the residual attribute inference of a frozen reference model with its utility assessment. Unlike PPO~\citep{schulman2017proximal}, GRPO estimates advantages from a group of sampled rewrites and needs no separately learned value network, which suits variable-length text generation and maps naturally onto comparing several candidate anonymizations of the same input. Because the policy is scored on its own samples, it explores rewrites the teacher never produced and pushes the privacy--utility frontier past the distilled ceiling. The same frozen model supplies the reward and anchors the update, so no external model is needed at training or inference.
Figure~\ref{fig:workflow} illustrates the full pipeline.

% Trained on Llama-3.1-8B~\citep{grattafiori2024llama}, \ours{} attains a stronger privacy--utility trade-off than the state-of-the-art DPO-distilled baseline and outperforms adversarial anonymization driven by frontier models such as Gemini~2.5~Flash~\citep{gemini25} and Claude~\citep{anthropic2025claude}, while running entirely on-device at roughly $1\%$ of the cost of the GPT-4o~\citep{achiam2023gpt} teacher from which it is distilled, and well below that of the frontier anonymizers. The advantage is consistent across three independent LLM judges and holds at every self-refinement step. In summary, our contributions are: \textbf{(1)} we cast local text anonymization as direct reinforcement learning against a self-generated adversarial reward, rather than offline imitation of a teacher; \textbf{(2)} we implement \ours{}, a GRPO recipe in which one small model serves as anonymizer, attacker, and utility judge; and \textbf{(3)} through extensive experiments and ablations we show that \ours{} improves over both distillation and frontier-model baselines on the privacy--utility trade-off while remaining cheap and fully local.

Trained on Llama-3.1-8B~\citep{grattafiori2024llama}, \ours{} attains
a stronger privacy--utility trade-off than the state-of-the-art
DPO-distilled baseline, consistently across three independent LLM
judges. Against adversarial anonymization driven by frontier models
such as Gemini~2.5~Flash~\citep{gemini25} and
Claude~\citep{anthropic2025claude}, it achieves a comparable or
better overall trade-off while removing substantially more private
information, all while running entirely on-device at roughly $1\%$
of the cost of the GPT-4o~\citep{achiam2023gpt} teacher from which it
is distilled, and well below that of the frontier anonymizers. In
summary, our contributions are: \textbf{(1)} we cast local text
anonymization as direct reinforcement learning against a
self-generated adversarial reward, rather than offline imitation of a
teacher; \textbf{(2)} we implement \ours{}, a GRPO recipe in which
one small model serves as anonymizer, attacker, and utility judge;
and \textbf{(3)} through extensive experiments and ablations we show
that \ours{} improves on the privacy--utility trade-off over
distillation, and over frontier-model baselines in privacy, while
remaining cheap and fully local.

% \section{Related Work}
% \input{src/relwork}

\section{Preliminaries}
\label{sec:related}
% Section 3: Preliminaries / Background.
% Shared foundation with SEAL (Staab et al. 2024; the SEAL framework): the
% inference threat model, adversarial anonymization, and the trajectory-based
% data used for distillation. Notation follows SEAL for consistency.
\label{s:pre}

We study anonymization against \emph{inference-time} privacy threats, in which a capable language model acts as an adversary that infers private attributes from seemingly innocuous text~\citep{staab2023beyond}. This section formalizes the threat model and reviews the adversarial anonymization procedure and the trajectory data on which our method builds. 

\paragraph{Threat model and problem statement.}
Let $x$ be a text written by a user that may leak private attributes $\mathcal{P}$ (e.g., age, gender, location, occupation) about its author. An adversary is a language model $\mathcal{M}_{\text{priv}}$ that, given $x$ and a target attribute set $\mathcal{P}$, predicts each attribute's value with a confidence score. Unlike classical identifiers removed by named-entity recognition or pattern matching, these attributes are often encoded in context-dependent semantic cues that survive surface-level scrubbing~\citep{staab2023beyond}. An anonymizer rewrites $x$ into $x'$ so that the attributes in $\mathcal{P}$ can no longer be reliably inferred from $x'$, while preserving the original semantic content and readability. The goal is thus a favorable \emph{privacy--utility trade-off}: strong protection against attribute inference at minimal cost to the meaning and fluency of the text.

\paragraph{Adversarial anonymization.}
A powerful way to obtain such rewrites is feedback-guided adversarial anonymization~\citep{staab2024language}, in which an anonymizer $\mathcal{M}_{\text{anon}}$ and an adversary $\mathcal{M}_{\text{priv}}$ alternate: the adversary infers the attributes that remain exposed, and the anonymizer edits the text to remove them. Given an initial text $x_0$ and target attributes $\mathcal{P}$, and additionally introducing a utility evaluator $\mathcal{M}_{\text{util}}$, the procedure repeats the following three steps for $t = 0, 1, \dots, T-1$: \textbf{(1) inference} $ \mathcal{P}_t \sim \mathcal{M}_{\text{priv}}(x_t, \mathcal{P})$, \textbf{(2) refinement} $x_{t+1} \sim \mathcal{M}_{\text{anon}}(x_t, \mathcal{P}_t)$, and \textbf{(3) utility} $\mathcal{U}_{t+1} \sim \mathcal{M}_{\text{util}}(x_0, x_{t+1}),$
% \begin{align*}
%     \text{(i) inference:}\quad & \mathcal{P}_t \sim \mathcal{M}_{\text{priv}}(x_t, \mathcal{P}), \\
%     \text{(ii) refinement:}\quad & x_{t+1} \sim \mathcal{M}_{\text{anon}}(x_t, \mathcal{P}_t), \\
%     \text{(iii) utility:}\quad & \mathcal{U}_{t+1} \sim \mathcal{M}_{\text{util}}(x_0, x_{t+1}),
% \end{align*}
where each inferred attribute $p \in \mathcal{P}_t$ carries a rationale and a confidence score $\text{conf}(p)$, and $\mathcal{U}_{t+1}$ collects utility measures (e.g., readability and semantic preservation) of $x_{t+1}$ relative to $x_0$. The procedure terminates after a fixed number of steps or once no further attributes can be inferred, yielding a trajectory
    $\tau = (s_0, s_1, \dots, s_T),$
where each state $s_i = (x_i, \mathcal{P}_i, \mathcal{U}_i)$ pairs a text with its inferred attributes and utility measurements. Running many LLMs as $\mathcal{M}_{\text{anon}}$, $\mathcal{M}_{\text{priv}}$, and $\mathcal{M}_{\text{util}}$ over a corpus produces a trajectory set $\mathcal{T}$ capturing how a text can be progressively anonymized~\citep{staab2024language}.

\paragraph{Scoring anonymizations.}
To compare states within a trajectory we use scalar privacy and utility scores. Following~\citet{staab2024language}, an anonymization is more private when fewer attributes are inferred and those that are inferred carry lower confidence, and it is more useful when its utility measures are higher on average:
\begin{gather}
    p(s_i) = \Big( -|\mathcal{P}_i|,\ -\tfrac{1}{|\mathcal{P}_i|}\!\!\sum_{m \in \mathcal{P}_i}\! \text{conf}(m) \Big), \\
    u(s_i) = \tfrac{1}{|\mathcal{U}_i|}\!\!\sum_{m \in \mathcal{U}_i}\! m .
\end{gather}
Privacy scores are ordered lexicographically (first by the number of inferred attributes, then by mean confidence), so that $p(s_j) > p(s_i)$ means $s_j$ is strictly more private than $s_i$; utility scores are compared by their average. The orderings identify, within each trajectory, which later rewrites dominate earlier ones in privacy and utility, and form the supervision signal for distilling anonymization behavior into a small language model.

\paragraph{Distillation setup.}
Our aim is to distill this behavior into a single small language model (SLM) that performs anonymization and its own evaluation, so that anonymization can run locally without querying external, potentially untrusted models~\citep{staab2024language}. Concretely, given the trajectory set $\mathcal{T}$ collected above, we train a target policy $\pi_\theta$ that (a) rewrites a text to improve its privacy--utility trade-off, (b) infers private attributes from a text as an adversary, and (c) evaluates the utility of a rewrite. Section~\ref{s:methods} describes how we adapt the model to these tasks and then optimize its anonymization policy with reinforcement learning.

\paragraph{Task adaptation via supervised fine-tuning.}
\label{ss:sft}
We adapt the base model to the three abilities for self-refinement: (i) \emph{anonymization}, rewriting a text to reduce attribute inferability; (ii) \emph{adversarial inference}, predicting private attributes with confidences; and (iii) \emph{utility evaluation}, judging semantic preservation and readability of a rewrite. Using the privacy and utility orderings defined above, we mine, from each trajectory $\tau \in \mathcal{T}$, the pairs in which a later rewrite dominates an earlier one, together with per-state inference and utility labels:
\begin{align*}
    \mathcal{D}_{\text{anon}} &= \{ (x_i, x_j) \mid 0 \le i < j \le T, \\
                              &\qquad\ \ p(s_j) > p(s_i),\ u(s_j) \ge u(s_i) \}, \\
    \mathcal{D}_{\text{priv}} &= \{ (x_i, \mathcal{P}_i) \mid s_i \in \tau \}, \\
    \mathcal{D}_{\text{util}} &= \{ (x_i, \mathcal{U}_i) \mid s_i \in \tau \}.
\end{align*}
The model $\pi_\theta$ is trained to minimize a weighted sum of next-token losses over the three datasets,
\begin{equation*}
\begin{aligned}
    \mathcal{L}_{\text{SFT}} = {}&
    \lambda_{\text{anon}}\, \mathcal{L}_{\text{anon}}(\mathcal{D}_{\text{anon}}) +
    \lambda_{\text{priv}}\, \mathcal{L}_{\text{priv}}(\mathcal{D}_{\text{priv}}) \\
    &+ \lambda_{\text{util}}\, \mathcal{L}_{\text{util}}(\mathcal{D}_{\text{util}}),
\end{aligned}
\end{equation*}
so that a single model can both generate and evaluate anonymizations. We denote the resulting model $\pi_{\text{ref}}$; it serves as the initialization for the reinforcement-learning stage and as the frozen reference used in its reward and policy regularizer.

\section{Proposed Algorithm}
\label{s:methods}

Our method trains a single small language model to be both a strong anonymizer and its own evaluator. We build on the supervised fine-tuned model $\pi_{\text{ref}}$ of Section~\ref{ss:sft}, which adapts the base model to anonymization, adversarial inference, and utility evaluation following the SEAL distillation recipe. Our central contribution is to replace the preference-learning (DPO) step of the prior distillation approach~\citep{kim2026self} with reinforcement learning that directly optimizes the privacy--utility trade-off. We fine-tune the anonymization policy with Group Relative Policy Optimization (GRPO)~\citep{shao2024deepseekmath} against an explicit reward, using the frozen reference model $\pi_{\text{ref}}$ itself as the adversary and utility evaluator.

\subsection{Policy Optimization for Anonymization via GRPO}
\label{ss:grpo}
Because SFT only imitates trajectory data, we fine-tune the anonymization policy with reinforcement learning, treating each rewrite as an action and scoring it with a reward that encodes both privacy and utility. Algorithm~\ref{alg:grasp} summarizes the procedure.

\paragraph{Reward design.}
Given an input text $x$ with target attribute set $\mathcal{P}$, the policy proposes an anonymization $x' \sim \pi_\theta(\cdot \mid x)$. We score $x'$ with two components. For \emph{privacy}, the reference model, acting as the adversary $\mathcal{M}_{\text{priv}}=\pi_{\text{ref}}$, attempts to infer the target attributes from $x'$; let $\mathcal{P}' \subseteq \mathcal{P}$ be the attributes for which it produces a confident guess, each with confidence $\text{conf}(m)$. We count \emph{any} confident inference, irrespective of whether the guess is correct (we return to this choice below). The privacy reward then rewards hiding attributes and lowering the adversary's confidence,
\begin{equation}
\label{eq:r-priv}
\begin{aligned}
    r_{\text{p}}(x') = {}& \alpha\Big(1 - \tfrac{|\mathcal{P}'|}{|\mathcal{P}|}\Big) \\
    & + \beta\Big(1 - \tfrac{1}{|\mathcal{P}'|}\!\!\sum_{m \in \mathcal{P}'}\! \text{conf}(m)\Big),
\end{aligned}
\end{equation}
with $\alpha + \beta = 1$ and the convention that the confidence term equals $1$ when $\mathcal{P}' = \emptyset$ (so a fully anonymized rewrite attains $r_{\text{p}}=1$). For \emph{utility}, the same reference model acts as the evaluator $\mathcal{M}_{\text{util}}=\pi_{\text{ref}}$, comparing $x'$ to the original $x$ and returning normalized scores for readability, semantic preservation, and freedom from hallucinated content, whose average we denote $r_{\text{u}}(x') \in [0,1]$. The two components are combined subject to two hard constraints that rule out degenerate rewrites:
\begin{equation}
\label{eq:reward}
    R(x') =
    \begin{cases}
        w_{\text{p}}\, r_{\text{p}}(x') + w_{\text{u}}\, r_{\text{u}}(x'), & x' \text{ feasible}, \\[2pt]
        -1, & \text{otherwise,}
    \end{cases}
\end{equation}
where a rewrite $x'$ is \emph{feasible} when it clears a utility floor, $r_{\text{u}}(x') \ge \tau$, and stays within a length band, $|x'| \in [\tfrac{1}{2}|x|,\, 2|x|]$ in words. The length constraint discourages trivial solutions that delete or pad the text, and the utility floor $\tau$ prevents the policy from purchasing privacy at the cost of destroying meaning; $w_{\text{p}}$ and $w_{\text{u}}$ set the privacy--utility balance.

\paragraph{Design choices against reward hacking.}
The challenging part is not switching from preference learning to GRPO, but making the on-policy optimization \textit{stable}. Because GRPO trains the policy on its own samples, the policy will exploit any weaknesses in the reward instead of learning the intended behavior. Each part of the reward addresses a specific failure we encountered during training.
\begin{itemize}[leftmargin=1.3em, itemsep=2pt, topsep=3pt]
    \item \textbf{Confidence makes the signal learnable.} Counting only how many attributes stay exposed is a coarse, almost all-or-nothing score, so rewrites in a group often tie, and GRPO finds no difference to learn from. The continuous confidence term rewards partial progress, separating a rewrite that merely lowers the adversary's certainty from one that removes a cue outright. Without it, the model stops hiding information and privacy falls back to nearly the original level (ablations are presented in Table~\ref{table:ablation}).
    
    \item   \textbf{Penalize any confident guess, not only correct ones.} A truth-aligned reward that only penalizes the adversary for being \emph{right} can be satisfied without removing information: the policy keeps the revealing cue but rewords it just enough to make the reference attacker guess wrong, leaving the private detail in place, only disguised. A stronger evaluation attacker sees through the disguise and recovers it, so real privacy does not improve (truth-aligned reward, Table~\ref{table:ablation}). Penalizing \emph{any} confident guess instead forces the policy to delete the cue.

    \item \textbf{The length band blocks empty rewrites.} With no length limit, the policy discovers that a very short, near-empty rewrite leaks nothing and scores high on privacy while destroying the content. Requiring the output to stay within $[\tfrac{1}{2}|x|, 2|x|]$ words removes this shortcut.
    
    \item \textbf{The utility floor keeps text usable.} Without a floor, the policy keeps trading meaning for privacy over successive self-refinement rounds and utility drops sharply. The floor $r_{\text{u}} \ge \tau$ keeps every rewrite readable and faithful, giving a better-balanced result ($\tau{=}0$, Table~\ref{table:ablation}).
\end{itemize}
\paragraph{Reference model as adversary.}
A key design choice is that both $\mathcal{M}_{\text{priv}}$ and $\mathcal{M}_{\text{util}}$ in the reward are instantiated by the frozen reference model $\pi_{\text{ref}}$ from Section~\ref{ss:sft}, rather than a separately trained attacker or critic. Because SFT already equips the model with adversarial-inference and utility-evaluation abilities, $\pi_{\text{ref}}$ supplies the reward, and the same frozen model serves as the KL anchor of the policy update below. This keeps training self-contained and avoids a second network.

\paragraph{Our GRPO objective.}
We optimize the policy with GRPO~\citep{shao2024deepseekmath}, which estimates advantages from \emph{groups} of sampled outputs and therefore requires no learned value network, unlike PPO~\citep{schulman2017proximal}. For each input $x$, we draw a group of $G$ anonymizations $\{x'_1, \dots, x'_G\} \sim \pi_{\theta_{\text{old}}}(\cdot \mid x)$, score each with $R_i = R(x'_i)$ from Eq.~\eqref{eq:reward}, and normalize the rewards within the group to form advantages
    $\hat{A}_i = \frac{R_i - \operatorname{mean}(\{R_j\}_{j=1}^{G})}{\operatorname{std}(\{R_j\}_{j=1}^{G})},$
which are shared across all tokens of $x'_i$. The policy is updated by maximizing the clipped surrogate objective with a KL penalty to the reference model,
\begin{equation}
\label{eq:grpo}
\begin{aligned}
    \mathcal{J}(\theta) &= \mathbb{E}\Big[ \tfrac{1}{G}\textstyle\sum_{i}\, \tfrac{1}{|x'_i|} \sum_{t}\, \ell_{i,t}(\theta) \Big] \\
    &\quad - \beta_{\text{KL}}\, \mathbb{D}_{\text{KL}}\!\big(\pi_\theta \,\|\, \pi_{\text{ref}}\big),
\end{aligned}
\end{equation}
where the outer average runs over the $G$ group samples and the inner over the tokens of $x'_i$, and the per-token clipped surrogate is
\begin{equation}
\label{eq:grpo-surrogate}
    \ell_{i,t}(\theta) = \min\!\big( \rho_{i,t}\,\hat{A}_i,\ \operatorname{clip}(\rho_{i,t}, 1{-}\epsilon, 1{+}\epsilon)\,\hat{A}_i \big).\nonumber
\end{equation}
Here $\epsilon$ is the clipping range, $\beta_{\text{KL}}$ the strength of the reference regularizer, and the token-level importance ratio is
    $\rho_{i,t} = \frac{\pi_\theta(x'_{i,t} \mid x, x'_{i,<t})}{\pi_{\theta_{\text{old}}}(x'_{i,t} \mid x, x'_{i,<t})}$.
 Intuitively, anonymizations that beat their group-mates on the privacy--utility reward are up-weighted, while the KL term keeps the policy close to the SFT reference and preserves fluency.

\subsection{Iterative Self-Refinement at Inference}
\label{ss:refine}
At test time the trained policy anonymizes by self-refinement, using no external model. Starting from $x_0$, at each step $t$ the model infers the residual private attributes $\mathcal{P}_t^{\pi}$ and evaluates the utility $\mathcal{U}_t^{\pi}$ of the current text, then produces a refined rewrite conditioned on its own feedback,
\begin{equation*}
    x_{t+1} \sim \pi_\theta(\cdot \mid x_t, \mathcal{P}_t^{\pi}, \mathcal{U}_t^{\pi}).
\end{equation*}
The loop continues until a target privacy--utility trade-off is reached or a fixed iteration budget is exhausted. Because a single local model performs both generation and critique, anonymization requires no proprietary or external adversary, and users can steer the number of refinement rounds to trade privacy against utility as they prefer.

\section{Empirical Results}
\label{s:results}

\subsection{Setup}
\label{ss:setup}

\paragraph{Datasets.}
% We build on SynthPAI~\citep{yukhymenko2024synthpai}, a corpus of synthetic personal profiles paired with text comments generated from those profiles. We use the 3{,}456 comments with high-quality human labels for eight personal attributes: age, education level, gender, income level, location, marital status, occupation, and place of birth. We use the anonymization trajectories of the SFT+DPO distillation baseline (SEAL). These trajectories come from simulating adversarial anonymization~\citep{staab2024language} for up to three steps, with GPT-4o~\citep{achiam2023gpt} acting as anonymizer, attribute-inference model, and utility evaluator. Of the 300 profiles, 275 are used for trajectory generation (2{,}734 comments) and 25 are held out for evaluation (723 comments), which we call the \emph{main} set. Following the same protocol, we additionally evaluate on the \emph{hard} set of 500 texts that embed personal information contextually rather than as explicit identifiers. These are constructed from the held-out profiles and never used for training, testing generalization to harder, unseen cases. This mirrors the intended use of the framework, in which trajectories are distilled from synthetic profiles using an external LLM while the resulting small model runs locally on real, private data without invoking an untrusted external service.

We evaluate on SynthPAI~\citep{yukhymenko2024synthpai}, a corpus of synthetic personal profiles paired with text comments labeled for eight personal attributes (age, education level, gender, income level, location, marital status, occupation, and place of birth). We use the anonymization trajectories of the SEAL distillation baseline, produced by adversarial anonymization~\citep{staab2024language} with GPT-4o~\citep{achiam2023gpt}. We evaluate on two held-out splits never seen in training: a \emph{main} set of $723$ comments and a \emph{hard} set of $500$ texts that embed personal information contextually rather than as explicit identifiers, testing generalization to harder, unseen cases. More details on the dataset can be found in Appendix~\ref{app:datasets}.

\paragraph{Baselines.}
We compare \ours{} against the original (unmodified) text and several anonymization methods. As a rule-based reference we include Azure's PII detection tool~\citep{microsoft2025}, which redacts sensitive spans via named-entity recognition. As a pure semantic-rewriting reference we include Dipper~\citep{krishna2023paraphrasing}, an 11B paraphraser not designed for anonymization. As the strongest prior LLM-based method we include adversarial anonymization~\citep{staab2024language}, which iteratively rewrites text through interaction between an anonymizer and an inference model, run with three frontier anonymizers (Gemini~2.5~Flash~\citep{gemini25}, Claude~Haiku~4.5~\citep{anthropic2025claude}, and Qwen3~\citep{qwen3blog}). Finally, we compare against SEAL~\cite{kim2026self}, the SFT+DPO distillation baseline, which is the most direct point of comparison for isolating the effect of reinforcement learning.

% \paragraph{Evaluating privacy and utility.}
% To measure privacy, we use GLM-5 as the inference model, given its strong attribute-inference ability~\citep{staab2023beyond}. The model performs zero-shot chain-of-thought inference over the eight attributes, and we report attribute-inference accuracy, the average fraction of correctly inferred attributes, so that lower is better. For utility, we follow prior work~\citep{staab2024language} and use a GLM-5 judge to score each anonymization on (1) readability, (2) semantic preservation relative to the input, and (3) absence of hallucinated content, and we take their average as the utility score. We summarize the trade-off with an \emph{overall} score, the relative privacy improvement minus the relative utility loss, normalized by the original privacy. To confirm that our conclusions are not an artifact of a single evaluator, we additionally repeat the assessment with two independent judges, Qwen3 and GPT-OSS (Section~\ref{ss:ablations}).

\paragraph{Evaluating privacy and utility.}
To measure privacy, we use GLM-5~\citep{glm2025} as the inference model. Since capable LLMs are strong attribute-inference attackers~\citep{staab2023beyond}, adopting a powerful model here makes the privacy evaluation stringent. The model performs zero-shot chain-of-thought inference over the eight attributes, and we report attribute-inference accuracy, the average fraction of correctly inferred attributes, so that lower is better. For utility, we use a GLM-5 judge to score each anonymization on (1) readability, (2) semantic preservation relative to the input, and (3) absence of hallucinated content, and we take their average as the utility score. Following SEAL, we summarize the trade-off with an \emph{overall} score, the relative privacy improvement minus the relative utility loss, normalized by the original privacy. To confirm that our conclusions are not an artifact of a single evaluator, we additionally repeat the assessment with two independent judges, Qwen3 and GPT-OSS~\citep{openai2025gptoss} (Section~\ref{ss:ablations}).

\begin{table*}[t]
    \centering
    \footnotesize
    \caption{\textbf{Anonymization on the main dataset}, judged by GLM-5. For SEAL and \ours{} we report the best-Overall ($_{\text{O}}$) and best-privacy ($_{\text{P}}$) rounds; full trajectories are in Table~\ref{table:main-table-iter}. \textbf{Bold}/\underline{underline}: best/second-best per row.}
    \label{table:main-table}
    \vspace{0.05in}
    \resizebox{.8\textwidth}{!}{%
    \begin{tabular}{l c cc ccc cc cc}
        \toprule[1pt]
        \multirow{2.5}{*}{\textbf{Metric}} & \multirow{2.5}{*}{\textbf{Original}} & \multirow{2.5}{*}{\textbf{Azure}} & \multirow{2.5}{*}{\textbf{Dipper}} & \multicolumn{3}{c}{\textbf{Adv. Anon.}} & \multicolumn{2}{c}{\textbf{SEAL}} & \multicolumn{2}{c}{\textbf{\ours{} (Ours)}} \\
        \cmidrule(lr){5-7} \cmidrule(lr){8-9} \cmidrule(lr){10-11}
         & & & & \scriptsize{Gemini} & \scriptsize{Claude} & \scriptsize{Qwen3} & \scriptsize{best-O} & \scriptsize{best-P} & \scriptsize{best-O} & \scriptsize{best-P} \\
        \midrule[0.75pt]
\rowcolor{RowHighlight} \textbf{Overall $\uparrow$} & - & -0.033 & -0.018 & 0.284 & 0.292 & 0.207 & 0.368 & 0.304 & \textbf{0.396} & \underline{0.368} \\
        \midrule[0.50pt]
        \rowcolor{RowHighlight} \textbf{Privacy $\downarrow$} & 0.620 & 0.610 & 0.538 & 0.367 & 0.365 & 0.454 & 0.262 & \underline{0.244} & 0.260 & \textbf{0.195} \\
        \midrule[0.25pt]
        \hspace{1em} Age & 0.762 & 0.802 & 0.614 & 0.720 & 0.614 & 0.750 & 0.551 & \underline{0.500} & 0.600 & \textbf{0.300} \\
        \hspace{1em} Edu & 0.645 & 0.667 & 0.718 & 0.521 & 0.510 & 0.576 & 0.416 & 0.325 & \underline{0.266} & \textbf{0.203} \\
        \hspace{1em} Gnd & 0.883 & 0.820 & 0.721 & 0.508 & 0.590 & 0.623 & 0.450 & \underline{0.417} & 0.417 & \textbf{0.250} \\
        \hspace{1em} Inc & 0.561 & 0.598 & 0.531 & 0.510 & \underline{0.490} & 0.561 & \textbf{0.457} & 0.521 & 0.607 & 0.536 \\
        \hspace{1em} Loc & 0.470 & 0.362 & 0.300 & 0.067 & 0.090 & 0.097 & \textbf{0.015} & \underline{0.023} & 0.050 & 0.050 \\
        \hspace{1em} Mar & 0.635 & 0.676 & 0.541 & 0.554 & 0.608 & 0.693 & \underline{0.324} & 0.425 & 0.409 & \textbf{0.182} \\
        \hspace{1em} Occ & 0.578 & 0.542 & 0.448 & 0.163 & 0.167 & 0.289 & 0.069 & \textbf{0.048} & \underline{0.069} & 0.083 \\
        \hspace{1em} PoB & 0.321 & 0.286 & 0.167 & 0.143 & 0.179 & 0.143 & \textbf{0.071} & \underline{0.125} & 0.500 & 0.500 \\
        \midrule[0.50pt]
        \rowcolor{RowHighlight} \textbf{Utility $\uparrow$} & 1.000 & \textbf{0.951} & 0.850 & 0.876 & 0.880 & \underline{0.940} & 0.791 & 0.698 & 0.815 & 0.683 \\
        \midrule[0.25pt]
        \hspace{1em} Mean & 1.000 & \textbf{0.936} & 0.848 & 0.750 & 0.755 & \underline{0.873} & 0.598 & 0.474 & 0.609 & 0.421 \\
        \hspace{1em} Read & 1.000 & 0.922 & 0.950 & 0.988 & 0.991 & \textbf{0.998} & 0.991 & 0.983 & \underline{0.993} & 0.982 \\
        \hspace{1em} Hall & 1.000 & \textbf{0.994} & 0.752 & 0.888 & 0.893 & \underline{0.947} & 0.783 & 0.636 & 0.844 & 0.646 \\
        \bottomrule[1pt]
    \end{tabular}%
    }
\end{table*}

\begin{table*}[t]
    \centering
    \footnotesize
    \caption{\textbf{Anonymization on the hard dataset}, judged by GLM-5. For SEAL and \ours{} we report the best-Overall ($_{\text{O}}$) and best-privacy ($_{\text{P}}$) rounds; full trajectories are in Table~\ref{table:hard-table-iter}. \textbf{Bold}/\underline{underline}: best/second-best per row.}
    \label{table:hard-table}
    \vspace{0.05in}
    \resizebox{0.8\textwidth}{!}{%
    \begin{tabular}{l c cc ccc cc cc}
        \toprule[1pt]
        \multirow{2.5}{*}{\textbf{Metric}} & \multirow{2.5}{*}{\textbf{Original}} & \multirow{2.5}{*}{\textbf{Azure}} & \multirow{2.5}{*}{\textbf{Dipper}} & \multicolumn{3}{c}{\textbf{Adv. Anon.}} & \multicolumn{2}{c}{\textbf{SEAL}} & \multicolumn{2}{c}{\textbf{\ours{} (Ours)}} \\
        \cmidrule(lr){5-7} \cmidrule(lr){8-9} \cmidrule(lr){10-11}
         & & & & \scriptsize{Gemini} & \scriptsize{Claude} & \scriptsize{Qwen3} & \scriptsize{best-O} & \scriptsize{best-P} & \scriptsize{best-O} & \scriptsize{best-P} \\
        \midrule[0.75pt]
\rowcolor{RowHighlight} \textbf{Overall $\uparrow$} & - & -0.023 & -0.035 & \textbf{0.301} & \underline{0.299} & 0.265 & 0.266 & 0.254 & 0.294 & 0.281 \\
        \midrule[0.50pt]
        \rowcolor{RowHighlight} \textbf{Privacy $\downarrow$} & 0.781 & 0.734 & 0.707 & 0.465 & 0.454 & 0.517 & 0.420 & \underline{0.369} & 0.379 & \textbf{0.332} \\
        \midrule[0.25pt]
        \hspace{1em} Age & 0.912 & 0.868 & 0.789 & 0.738 & 0.717 & 0.800 & 0.655 & \underline{0.589} & 0.608 & \textbf{0.534} \\
        \hspace{1em} Edu & 0.834 & 0.802 & 0.828 & 0.665 & 0.674 & 0.744 & 0.616 & 0.512 & \underline{0.501} & \textbf{0.417} \\
        \hspace{1em} Gnd & 1.000 & 0.838 & 0.952 & 0.738 & \underline{0.571} & 0.595 & 0.667 & 0.619 & 0.571 & \textbf{0.500} \\
        \hspace{1em} Inc & 0.520 & 0.541 & 0.484 & 0.520 & 0.529 & 0.573 & 0.508 & \underline{0.467} & 0.487 & \textbf{0.457} \\
        \hspace{1em} Loc & 0.891 & 0.796 & 0.761 & 0.219 & 0.217 & 0.227 & \textbf{0.203} & \underline{0.203} & 0.212 & 0.209 \\
        \hspace{1em} Mar & 0.907 & 0.826 & 0.840 & 0.662 & 0.680 & 0.733 & 0.667 & \underline{0.613} & 0.627 & \textbf{0.527} \\
        \hspace{1em} Occ & 0.624 & 0.566 & 0.555 & 0.291 & 0.265 & 0.388 & 0.214 & \underline{0.147} & 0.169 & \textbf{0.117} \\
        \hspace{1em} PoB & 0.908 & 0.831 & 0.806 & 0.155 & \textbf{0.118} & 0.168 & 0.126 & 0.124 & 0.126 & \underline{0.120} \\
        \midrule[0.50pt]
        \rowcolor{RowHighlight} \textbf{Utility $\uparrow$} & 1.000 & \underline{0.916} & 0.870 & 0.896 & 0.879 & \textbf{0.927} & 0.804 & 0.727 & 0.780 & 0.706 \\
        \midrule[0.25pt]
        \hspace{1em} Mean & 1.000 & \textbf{0.898} & \underline{0.893} & 0.722 & 0.678 & 0.811 & 0.572 & 0.449 & 0.483 & 0.360 \\
        \hspace{1em} Read & 1.000 & 0.850 & 0.970 & \underline{0.995} & 0.995 & \textbf{0.999} & 0.993 & 0.981 & 0.966 & 0.947 \\
        \hspace{1em} Hall & 1.000 & \textbf{1.000} & 0.748 & \underline{0.972} & 0.966 & 0.972 & 0.848 & 0.750 & 0.890 & 0.811 \\
        \bottomrule[1pt]
    \end{tabular}%
    }
\end{table*}

\begin{figure*}[t]
    \centering
    \begin{subfigure}{0.45\textwidth}
        \centering
        \includegraphics[width=\linewidth]{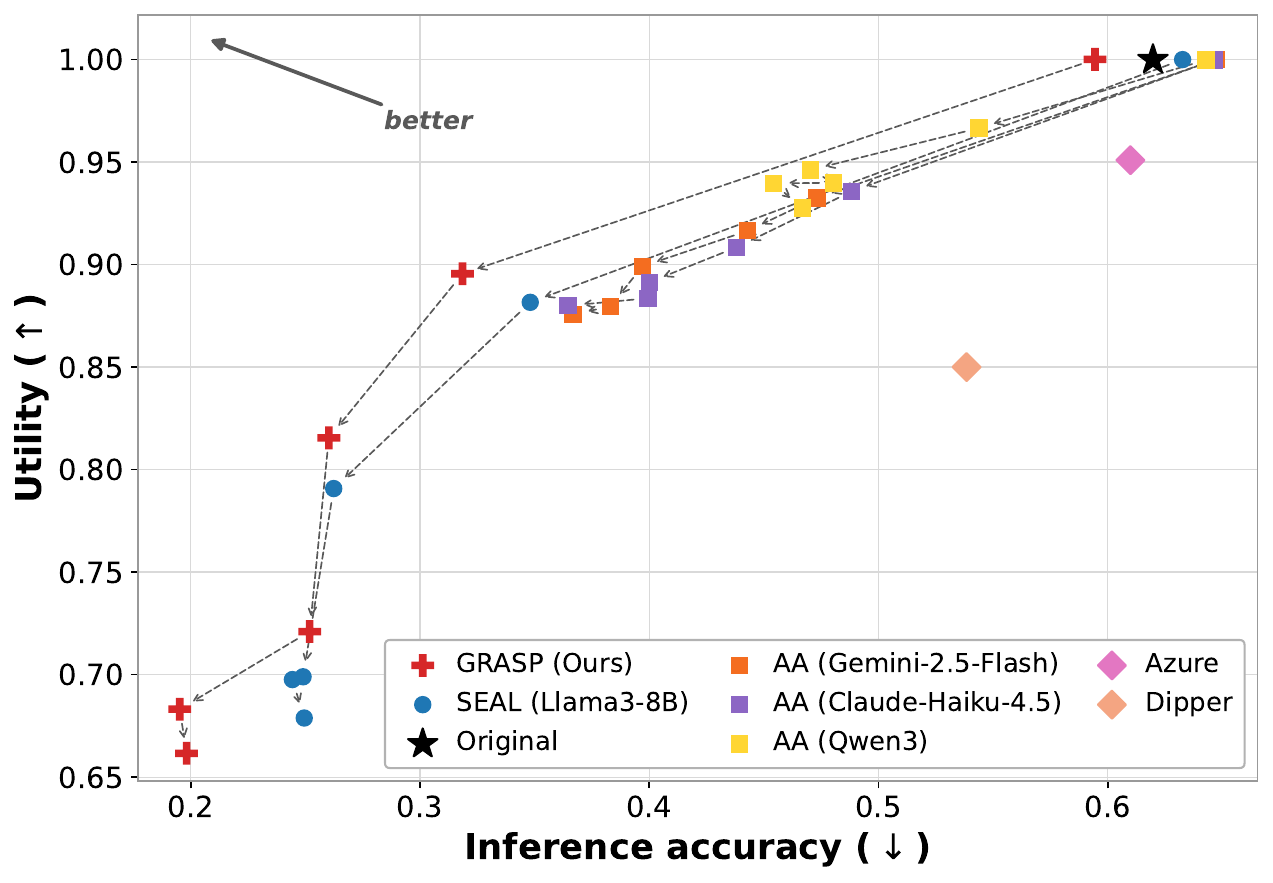}
        \caption{Main dataset}
        \label{fig:tradeoff-main}
    \end{subfigure}
    \hspace{2pt}%
    \begin{subfigure}{0.45\textwidth}
        \centering
        \includegraphics[width=\linewidth]{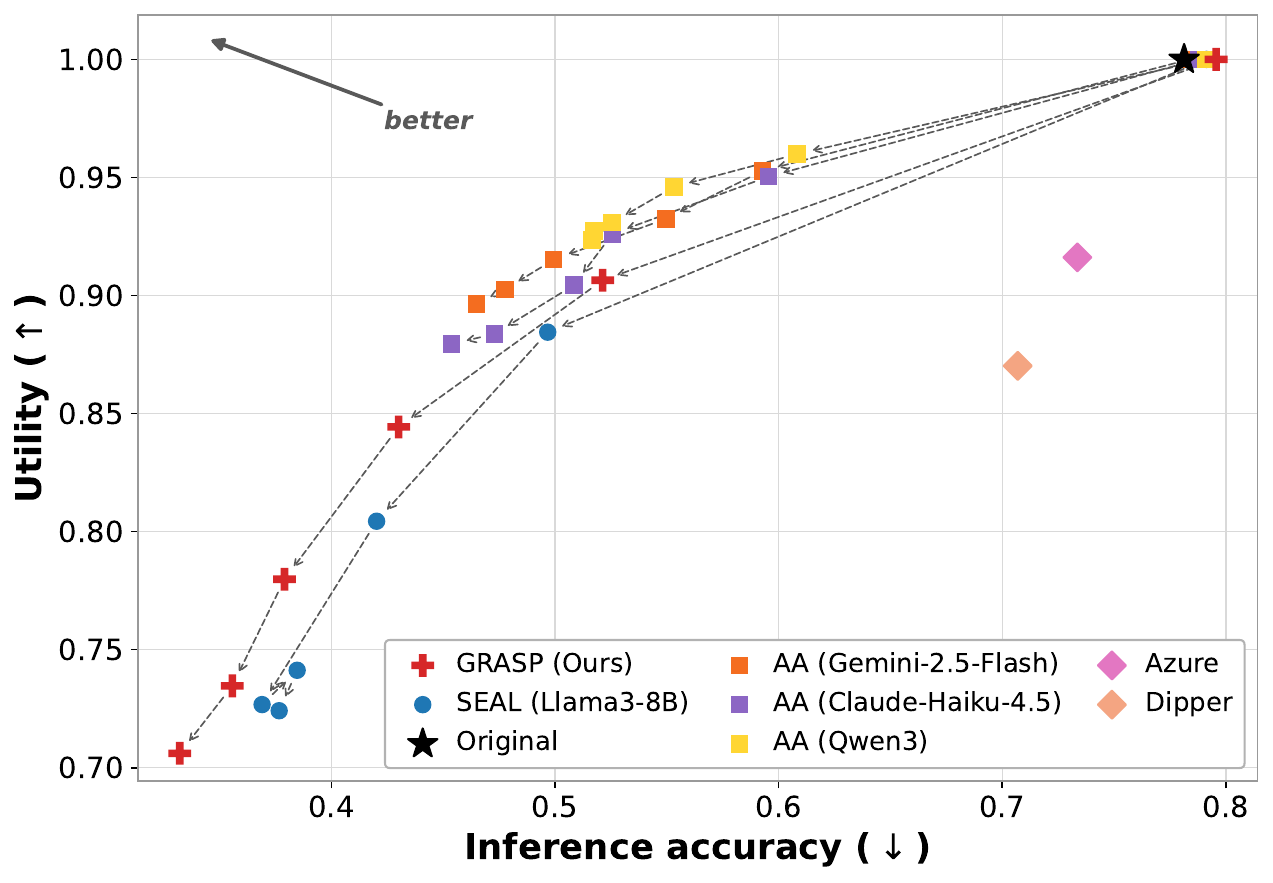}
        \caption{Hard dataset}
        \label{fig:tradeoff-hard}
    \end{subfigure}
    \caption{\textbf{Privacy--utility trade-off.} Self-refinement trajectories (rounds R0--R5) judged by GLM-5; the upper-left corner is ideal. \ours{} extends the Pareto frontier on both splits,
reaching the strongest anonymization.}
    \label{fig:tradeoff}
\end{figure*}

\paragraph{Implementation details.}
% Unless noted otherwise, \ours{} is trained on Llama-3.1-8B-Instruct~\citep{grattafiori2024llama}, and we also report results on Qwen3-8B~\citep{qwen3blog} to test generalization across model families. The SFT stage follows the distillation baseline exactly, using LoRA with rank $16$, $\alpha{=}16$, and dropout $0.05$, AdamW at learning rate $2\mathrm{e}{-}4$, and one epoch. We then initialize the policy from the SFT model, which also serves as the frozen reference $\pi_{\text{ref}}$, and optimize it with GRPO for a single refinement round on $5{,}000$ training prompts. Each prompt draws $G{=}8$ sampled anonymizations, and training uses learning rate $5\mathrm{e}{-}6$, KL weight $\beta_{\text{KL}}{=}0.03$, effective batch size $32$, sampling temperature $0.9$, and a maximum completion length of $512$ tokens. The reward uses privacy coefficients $\alpha{=}0.7,\beta{=}0.3$, trade-off weights $w_{\text{p}}{=}0.9,w_{\text{u}}{=}0.1$, and utility floor $\tau{=}0.55$. All models are trained with LoRA in mixed precision with FlashAttention-2 on $4{\times}$NVIDIA~A6000 (46\,GB) GPUs. At inference the trained policy anonymizes by self-refinement for up to five iterations. Full hyperparameters are given in the Appendix.

We train \ours{} on Llama-3.1-8B-Instruct~\citep{grattafiori2024llama}. The SFT stage follows the distillation baseline exactly, using LoRA with rank $16$, $\alpha{=}16$, and dropout $0.05$, AdamW at learning rate $2\mathrm{e}{-}4$, and one epoch. We then initialize the policy from the SFT model, which also serves as the frozen reference $\pi_{\text{ref}}$, and optimize it with GRPO for one epoch on $5{,}000$ training prompts. Each prompt draws $G{=}8$ sampled anonymizations, and training uses learning rate $5\mathrm{e}{-}6$, KL weight $\beta_{\text{KL}}{=}0.03$, effective batch size $32$, sampling temperature $0.9$, and a maximum completion length of $512$ tokens. The reward uses privacy coefficients $\alpha{=}0.7,\beta{=}0.3$, trade-off weights $w_{\text{p}}{=}0.9,w_{\text{u}}{=}0.1$, and utility floor $\tau{=}0.55$. All models are trained with LoRA~\citep{hu2022lora} on NVIDIA GPUs. At inference the trained policy anonymizes by self-refinement for up to five iterations. Table~\ref{table:hyperparams} summarizes all hyperparameters for both stages.

\subsection{Overall Performance}
\label{ss:main-results}

\paragraph{Main dataset.}
% Table~\ref{table:main-table} reports results on the main dataset for a Llama-3.1-8B model. Among the baselines, adversarial anonymization is by far the strongest, with Claude attaining the best trade-off (overall $0.292$), followed by Gemini ($0.284$) and Qwen3 ($0.207$), while the rule-based Azure and the paraphrasing Dipper are ineffective and even push the overall score below zero. \ours{} clearly surpasses all of them. With a \emph{single} refinement step it already reaches an overall of $0.382$, above the best iterate of every baseline, including SEAL, and its best round attains an overall of $\mathbf{0.396}$ at attribute-inference accuracy $0.260$, a markedly better privacy level than the strongest frontier baseline ($0.365$) for a modest utility cost. Relative to the SFT+DPO baseline, \ours{} improves both the overall score ($0.396$ vs.\ $0.368$) and the strongest achievable privacy. As self-refinement continues, privacy improves further, driving inference accuracy down to $0.195$, the lowest of any method, while readability stays above $0.98$. The gains are broad across attributes, with the largest reductions for location, occupation, and education level. Azure and Dipper, by contrast, yield minimal privacy improvement, and Dipper incurs the largest utility loss.

Table~\ref{table:main-table} reports results on the main dataset.
Among the baselines, SEAL is the strongest
(overall $0.368$), ahead of adversarial anonymization with frontier
models (Claude $0.292$, Gemini $0.284$, Qwen3 $0.207$), while the
rule-based Azure and the paraphrasing Dipper barely reduce attribute
inference and push the overall score below zero. \ours{} improves on
all of them. At its best-Overall round it reaches $\mathbf{0.396}$,
reducing attribute-inference accuracy to $0.260$---a $29\%$ relative
reduction from the best frontier anonymizer (Claude, $0.365$)---at a
$7\%$ relative utility cost ($0.815$ vs.\ $0.880$). With continued
self-refinement, \ours{} reaches inference accuracy $0.195$, below
SEAL's most private round ($0.244$) and the lowest of any method,
while keeping readability at $0.982$. Its advantage over SEAL is
concentrated on contextually encoded attributes: at the most private
round it lowers residual inference on age ($0.300$ vs.\ $0.500$),
marital status ($0.182$ vs.\ $0.425$), gender ($0.250$ vs.\ $0.417$),
and education ($0.203$ vs.\ $0.325$), while both methods drive
location and occupation close to zero. As Fig.~\ref{fig:tradeoff}
shows, \ours{} traces the outer privacy--utility frontier on this
split, whereas Azure and Dipper yield little privacy gain and Dipper
incurs the largest utility loss.

\paragraph{Hard dataset.}
On the harder set of texts with contextually embedded personal
information, all methods remove fewer attributes
(Table~\ref{table:hard-table}). The frontier
adversarial-anonymization models are more competitive on the
aggregate score, with Gemini ($0.301$) and Claude ($0.299$) narrowly
ahead of \ours{}'s best overall of $0.294$. This parity, however,
reflects a different operating point rather than comparable
anonymization: the frontier models preserve utility by stopping far
short on privacy, leaving inference accuracy between $0.454$ and
$0.517$, whereas \ours{} continues to reduce it to $0.332$ with
further refinement, the lowest of any method on this split. \ours{}
also improves on the SEAL baseline ($0.294$ vs.\ $0.266$ overall); as
on the main split, its advantage over SEAL is largest on education
($0.417$ vs.\ $0.512$ at the most private round), gender ($0.500$
vs.\ $0.619$), marital status ($0.527$ vs.\ $0.613$), and occupation
($0.117$ vs.\ $0.147$). Utility remains comparable to the main
split, with readability above $0.93$ at every round. As
Fig.~\ref{fig:tradeoff} shows, \ours{} extends the privacy--utility
frontier on this split as well, reaching inference accuracies no
baseline attains at any utility level.

\subsection{Ablations and Analysis}
\label{ss:ablations}
\vspace{-2pt}
\paragraph{Component ablation.}
Table~\ref{table:ablation} isolates the contribution of each design
choice, starting from the supervised fine-tuned model and toggling
one component of the GRPO stage at a time. All variants are evaluated
over the first three self-refinement rounds (R0--R3), so that every
configuration is compared under an identical protocol; the full model
reaches privacy $0.234$ on the main split and $0.377$ on the hard
split. The most important component is the confidence term in the
privacy reward: without it, privacy collapses to $0.626$/$0.773$
(main/hard), close to the un-anonymized text ($0.620$/$0.781$),
since the reward no longer penalizes leaving an attribute exposed as
long as the attacker is not fully certain. The truth-aligned reward,
which penalizes only correct guesses, leaves main-split privacy
nearly intact ($0.252$) but degrades the hard split to $0.430$: the
policy learns to disguise cues just enough to mislead the reference
attacker, and the stronger evaluation attacker recovers them.
Removing the attacker feedback that conditions each rewrite
($0.294$/$0.387$), halving the training data ($0.253$/$0.426$), and
shrinking the GRPO group to $G{=}4$ ($0.333$/$0.506$) each weaken
privacy on both splits, confirming that self-critique, more
experience, and richer group comparisons all contribute. Removing
the utility floor ($\tau{=}0$) is the one variant that can improve
privacy ($0.313$ on hard), but it does so by sacrificing meaning:
utility falls with every refinement round
($0.815{\to}0.650$ main, $0.832{\to}0.677$ hard), the degenerate
trade the floor is designed to prevent. Full per-iteration results
for all variants are reported in Table~\ref{table:ablation-iter}.

\begin{table}[t]
    \centering
    \small
    \caption{\textbf{Ablation of \ours{} components}, reporting privacy
($\downarrow$) and utility ($\uparrow$) at the best-Overall round.}
    \label{table:ablation}
    \setlength{\tabcolsep}{3pt}
    \resizebox{0.95\linewidth}{!}{%
    \begin{tabular}{l cc cc}
        \toprule[1pt]
        \multirow{2.5}{*}{\textbf{Method}} & \multicolumn{2}{c}{\textbf{Main}} & \multicolumn{2}{c}{\textbf{Hard}} \\
        \cmidrule(lr){2-3} \cmidrule(lr){4-5}
        & Priv.\,$\downarrow$ & Util.\,$\uparrow$ & Priv.\,$\downarrow$ & Util.\,$\uparrow$ \\
        \midrule[0.75pt]
        SFT (no RL) & 0.388 & 0.905 & 0.564 & 0.917 \\
        \midrule[0.25pt]
        \rowcolor{RowHighlight} \textbf{\ours{} (full)} & 0.234 & 0.730 & 0.377 & 0.773 \\
        \hspace{1em} w/o confidence reward & 0.626 & 0.992 & 0.773 & 0.996 \\
        \hspace{1em} w/o attacker feedback & 0.294 & 0.823 & 0.387 & 0.761 \\
        \hspace{1em} w/ truth-aligned reward & 0.252 & 0.753 & 0.430 & 0.835 \\
        \hspace{1em} w/o utility floor ($\tau{=}0$) & 0.272 & 0.815 & 0.313 & 0.677 \\
        \hspace{1em} fewer samples ($2$k vs $5$k) & 0.253 & 0.749 & 0.426 & 0.851 \\
        \hspace{1em} smaller group ($G{=}4$) & 0.333 & 0.870 & 0.506 & 0.888 \\
        \bottomrule[1pt]
    \end{tabular}
    }
\end{table}

\paragraph{Evaluation with alternative LLM judges.}
To test whether our conclusions depend on the evaluator, we re-score
both splits with two judges from different model families, Qwen3 and
GPT-OSS, alongside our primary GLM-5 judge
(Table~\ref{table:judge-robustness}). \ours{} attains a higher best
overall score than SEAL in all six
judge$\times$split cells: from $0.342$ to $0.400$ (main) and $0.300$
to $0.346$ (hard) under Qwen3, and from $0.291$ to $0.319$ and
$0.299$ to $0.357$ under GPT-OSS. Absolute values shift between
judges, but the ranking is unchanged, and \ours{} leads SEAL at
nearly every refinement step under every judge (Table~\ref{table:judge-robustness-iter}), trailing only before
self-refinement takes effect on the hard split.

\begin{table}[t]
    \centering
    \small
    \caption{\textbf{Judge robustness.} Overall ($\uparrow$), privacy
($\downarrow$), and utility ($\uparrow$) at each method's best-Overall round under three judges. \textbf{Bold}: better per
metric.}
    \label{table:judge-robustness}
    \setlength{\tabcolsep}{3pt}
    \resizebox{0.95\linewidth}{!}{%
    \begin{tabular}{ll ccc ccc}
        \toprule[1pt]
        \multirow{2.5}{*}{\textbf{Judge}} & \multirow{2.5}{*}{\textbf{Split}} & \multicolumn{3}{c}{\textbf{SEAL}} & \multicolumn{3}{c}{\textbf{\ours{} (Ours)}} \\
        \cmidrule(lr){3-5} \cmidrule(lr){6-8}
         & & Ovr. & Priv. & Util. & Ovr. & Priv. & Util. \\
        \midrule[0.75pt]
        \multirow{2}{*}{GLM-5} & main & 0.368 & 0.262 & 0.791 & \textbf{0.396} & \textbf{0.260} & \textbf{0.815} \\
         & hard & 0.266 & 0.420 & \textbf{0.804} & \textbf{0.294} & \textbf{0.379} & 0.780 \\
        \midrule
        \multirow{2}{*}{Qwen3} & main & 0.342 & \textbf{0.236} & 0.723 & \textbf{0.400} & 0.258 & \textbf{0.817} \\
         & hard & 0.300 & 0.378 & \textbf{0.784} & \textbf{0.346} & \textbf{0.293} & 0.721 \\
        \midrule
        \multirow{2}{*}{GPT-OSS} & main & 0.291 & 0.317 & 0.802 & \textbf{0.319} & \textbf{0.309} & \textbf{0.817} \\
         & hard & 0.299 & \textbf{0.254} & 0.624 & \textbf{0.357} & 0.318 & \textbf{0.765} \\
        \bottomrule[1pt]
    \end{tabular}
    }
\end{table}

\paragraph{Efficiency.}
Because anonymization runs entirely on a local 8B model, no text
leaves the user's machine. At a 1:1 input--output ratio, the
Llama-3.1-8B backbone costs roughly $1\%$ of the GPT-4o teacher,
versus $30\%$ for Claude Haiku 4.5 and $14\%$ for Gemini 2.5 Flash.
Per-sample rewriting takes $0.64$s, versus $0.49$--$0.81$s for the
API systems; a full self-refinement step, which adds local attribute
inference on the rewrite, takes $13.8$s, versus $3.4$--$5.6$s. Full
measurements are in Table~\ref{table:efficiency} in the Appendix.

\paragraph{Qualitative comparison.}
Appendix~\ref{qualitative} traces how each method rewrites the
same low-income comment: frontier anonymizers soften the wording
but leave the income cue intact, SEAL over-generalizes into stilted
phrasing, and \ours{} removes the cue while keeping a natural,
coherent sentence.

\section{Conclusion}
We introduced \ours{}, which replaces the offline preference learning
of prior anonymizer distillation with online reinforcement learning:
a single small model serves as anonymizer, adversary, and utility
judge, and GRPO optimizes the privacy--utility objective directly
against this self-generated reward, with constraints that guard
against reward hacking. On SynthPAI, \ours{} improves the
privacy--utility trade-off over SEAL under three
independent judges, removes more private information than adversarial
anonymization driven by frontier models, and extends the Pareto
frontier on both evaluation splits---while running entirely on-device
at roughly $1\%$ of the teacher's cost. 

\clearpage

\section*{Limitations}
\label{sec:limitations}
\paragraph{Inference latency.}
A full self-refinement step runs attribute inference on the local
$8$B model, taking $13.8$s per sample versus $3.4$--$5.6$s for
API-based adversarial anonymization, which offloads this step to a
frontier model (Appendix~\ref{sec:eff}). This is the price of
keeping all computation on-device; plain rewriting is comparable to
the API systems ($0.64$s), and users who need lower latency can run
fewer refinement rounds at a reduced privacy level.

% \paragraph{Uneven per-attribute behavior.}
% While \ours{} lowers aggregate inference accuracy below every
% baseline, gains are not uniform across attributes: rarely inferable
% attributes such as place of birth have small support in the main
% split and their per-attribute estimates are correspondingly noisy,
% and income remains difficult for all methods. Aggregate metrics
% should therefore be read alongside the per-attribute results in
% Tables~\ref{table:main-table} and~\ref{table:hard-table}.

% \paragraph{Synthetic evaluation data.}
% Following prior work, we evaluate on SynthPAI, whose texts and
% ground-truth attributes are synthetic. This enables training and
% evaluation without exposing real personal data---consistent with the
% deployment setting our method targets---but transfer to real user
% text, other languages, and attribute sets beyond the eight studied
% here remains to be validated.

% \paragraph{LLM-based evaluation.}
% Privacy and utility are scored by LLM judges. We mitigate
% evaluator-specific artifacts by replicating all comparisons under
% three judges from different model families
% (Table~\ref{table:judge-robustness}), but LLM judges remain imperfect proxies,
% particularly for utility, where the judgment is open-ended rather than checkable against a reference \citep{ghiasvand2026can}.

\paragraph{LLM-based evaluation.} Privacy and utility are scored by LLM judges. Judge behavior is known to be sensitive to the evaluation protocol itself, including position and verbosity effects, and to vary with the task being scored \citep{hariri2026testtime, zheng2023judging}. We mitigate evaluator-specific artifacts by replicating all comparisons under three judges from different model families (Table 4), but LLM judges remain imperfect proxies, particularly for utility, where the judgment is open-ended rather than checkable against a reference~\citep{ghiasvand2026can}.

\paragraph{Fixed attribute set.}
\ours{} is trained and evaluated on the eight personal attributes
annotated in SynthPAI; protecting a different or broader attribute
set (e.g., health status or political views) would require
regenerating trajectories and retraining, and performance on
attributes outside this set is untested.

\paragraph{Potential risks.}
\ours{} reduces but does not eliminate attribute leakage, and its
privacy measurements reflect current inference models; users should
not treat its output as a guarantee of anonymity, particularly for
high-stakes disclosures. Conversely, because the model is trained to
predict private attributes as part of its self-critique, it could in
principle be misused as an inference attacker, although its
capability is distilled from, and remains below, that of the
publicly available frontier models already capable of such
inference.

\bibliography{main.bib}

\clearpage
\appendix

% \clearpage

% \clearpage

% \section{Example Appendix}
% \label{sec:appendix}

\begin{algorithm*}[t]
\caption{\ours{}}
\label{alg:grasp}
\begin{algorithmic}[1]
\Require reference model $\pi_{\text{ref}}$ (SFT, \S\ref{ss:sft}); group size $G$; weights $w_{\text{p}},w_{\text{u}}$; coefficients $\alpha,\beta$; utility floor $\tau$; KL weight $\beta_{\text{KL}}$
\Ensure anonymization policy $\pi_\theta$
\State Initialize $\pi_\theta \gets \pi_{\text{ref}}$
\While{not converged}
    \State Sample a text $x$ with target attribute set $\mathcal{P}$
    \State Draw a group $\{x'_1,\dots,x'_G\} \sim \pi_{\theta_{\text{old}}}(\cdot \mid x)$
    \For{$i = 1$ \textbf{to} $G$}
        \State $\mathcal{P}' \gets \mathcal{M}_{\text{priv}}(x'_i)$ \Comment{$\pi_{\text{ref}}$ as adversary}
        \State $r_{\text{p}} \gets \alpha(1{-}\tfrac{|\mathcal{P}'|}{|\mathcal{P}|}) + \beta(1{-}\overline{\mathrm{conf}})$ \Comment{Eq.~\eqref{eq:r-priv}}
        \State $r_{\text{u}} \gets \mathcal{M}_{\text{util}}(x'_i, x)$ \Comment{$\pi_{\text{ref}}$ as evaluator}
        \If{$r_{\text{u}} \ge \tau$ \textbf{and} $|x'_i| \in [\tfrac{1}{2}|x|, 2|x|]$}
            \State $R_i \gets w_{\text{p}}\, r_{\text{p}} + w_{\text{u}}\, r_{\text{u}}$
        \Else
            \State $R_i \gets -1$ \Comment{infeasible rewrite}
        \EndIf
    \EndFor
    \State $\hat{A}_i \gets \big(R_i - \operatorname{mean}(\{R_j\})\big) / \operatorname{std}(\{R_j\})$ for all $i$
    \State Update $\theta$ by maximizing $\mathcal{J}(\theta)$ with KL to $\pi_{\text{ref}}$ \Comment{Eq.~\eqref{eq:grpo}}
\EndWhile
\State \Return $\pi_\theta$
\end{algorithmic}
\end{algorithm*}

\section{Dataset Details}
\label{app:datasets}

We build on SynthPAI~\citep{yukhymenko2024synthpai}, a corpus of
synthetic personal profiles paired with text comments generated from
those profiles. We use the 3{,}456 comments with high-quality human
labels for eight personal attributes: age, education level, gender,
income level, location, marital status, occupation, and place of
birth. Because our contribution is the reinforcement-learning
objective rather than the distillation data, we reuse the
anonymization trajectories of the SFT+DPO distillation baseline
(SEAL)~\citep{kim2026self}. These trajectories come from simulating
adversarial anonymization~\citep{staab2024language} for up to three
steps, with GPT-4o~\citep{achiam2023gpt} acting as anonymizer,
attribute-inference model, and utility evaluator. Of the 300
profiles, 275 are used for trajectory generation (2{,}734 comments)
and 25 are held out entirely from training.

The held-out profiles yield two evaluation splits. The \emph{main}
set consists of the 723 comments written from the held-out profiles.
The \emph{hard} set consists of 500 texts that embed personal
information contextually rather than as explicit identifiers,
constructed by SEAL~\citep{kim2026self}: hard-to-anonymize texts are
first identified in the original SynthPAI pool via iterative
adversarial anonymization, and an LLM is then prompted to generate
20 new texts per held-out profile emulating their
characteristics---attributes conveyed through narrative and
distributed linguistic cues rather than replaceable keywords. Its
higher baseline inference accuracy ($0.781$ vs.\ $0.620$ on the main
set) reflects this increased difficulty.

This mirrors the intended use of the framework, in which
trajectories are distilled from synthetic profiles using an external
LLM, while the resulting small model runs locally on real, private
data without invoking an untrusted external service.

\begin{table}[t]
    \centering
    \small
    \caption{\textbf{Hyperparameters for the SFT and GRPO stages of \ours{}.} The SFT stage matches SEAL; GRPO is run for a single self-refinement round.}
    \label{table:hyperparams}
    \setlength{\tabcolsep}{3pt}
    \resizebox{0.95\linewidth}{!}{%
    \vspace{0.05in}
    \begin{tabular}{l|l|c|c}
        \toprule[1pt]
        & Parameter & SFT & GRPO \\
        \midrule[0.75pt]
        \multirow{5}{*}{Training} & Effective batch size & 16 & 32 \\
        & Train epochs / rounds & 1 & 1 \\
        & KL weight $\beta_{\text{KL}}$ & - & 0.03 \\
        & Sampling temperature & - & 0.9 \\
        & Mixed precision & bf16 & bf16 \\
        \midrule[0.50pt]
        \multirow{6}{*}{Optimization} & Optimizer & AdamW & AdamW \\
        & Learning rate & 2e-4 & 5e-6 \\
        & Weight decay & 1e-2 & 1e-2 \\
        & $(\beta_1,\beta_2)$ & (0.9, 0.999) & (0.9, 0.999) \\
        & Max gradient norm & 1.0 & 1.0 \\
        & Max completion length & - & 512 \\
        \midrule[0.50pt]
        \multirow{3}{*}{LoRA} & Rank & 16 & 16 \\
        & Alpha & 16 & 16 \\
        & Dropout & 0.05 & 0.05 \\
        \midrule[0.50pt]
        \multirow{5}{*}{Reward} & Group size $G$ & - & 8 \\
        & Privacy coeffs $(\alpha,\beta)$ & - & (0.7, 0.3) \\
        & Weights $(w_{\text{p}}, w_{\text{u}})$ & - & (0.9, 0.1) \\
        & Utility floor $\tau$ & - & 0.55 \\
        & Training prompts & - & 5{,}000 \\
        \bottomrule[1pt]
    \end{tabular}
    }
\end{table}

\section{Implementation Details}
\label{app:hyperparams}
Unless noted otherwise, \ours{} is trained on
Llama-3.1-8B-Instruct~\citep{grattafiori2024llama}. The SFT stage
follows the distillation baseline exactly, using LoRA (rank $16$,
$\alpha_{\text{LoRA}}{=}16$, dropout $0.05$), AdamW at learning rate
$2\mathrm{e}{-}4$, and one epoch. We then initialize the policy from
the SFT model, which also serves as the frozen reference
$\pi_{\text{ref}}$, and optimize it with GRPO for one epoch on
$5{,}000$ training prompts drawn from the trajectory states. Each
prompt draws $G{=}8$ sampled anonymizations, and training uses
learning rate $5\mathrm{e}{-}6$, KL weight
$\beta_{\text{KL}}{=}0.03$, effective batch size $32$, sampling
temperature $0.9$, and a maximum completion length of $512$ tokens.
The reward uses privacy coefficients $\alpha{=}0.7$, $\beta{=}0.3$,
trade-off weights $w_{\text{p}}{=}0.9$, $w_{\text{u}}{=}0.1$, and
utility floor $\tau{=}0.55$. All models are trained with LoRA in
mixed precision with FlashAttention-2 on NVIDIA GPUs. At inference the trained policy anonymizes by
self-refinement for up to five iterations.
Table~\ref{table:hyperparams} summarizes all hyperparameters for
both stages.

\section{Efficiency}
\label{sec:eff}
Table~\ref{table:efficiency} reports inference cost and per-sample
latency, measured over $50$ samples with \ours{} running on a single
NVIDIA GPU and the baselines accessed via their APIs. Cost
assumes a $1{:}1$ input-to-output token ratio and is stated relative
to the GPT-4o teacher: the Llama-3.1-8B backbone used by \ours{}
comes to roughly $1\%$, versus $30\%$ for Claude~Haiku~4.5 and
$14\%$ for Gemini~2.5~Flash. For rewriting alone, per-sample latency
is $0.64$s, within the range of the API models
($0.49$--$0.81$s). A full self-refinement step additionally runs
attribute inference on the rewrite; because \ours{} performs this
step with the same local $8$B model whereas the baselines use their
frontier API model, its latency rises to $13.8$s versus
$3.4$--$5.6$s. Note that the latency comparison is approximate, as
the API models run on optimized serving infrastructure while \ours{}
runs on a single local GPU; the corresponding trade-off is discussed
in the Limitations Section.

\begin{table}[t]
    \centering
    \small
    \caption{\textbf{Efficiency.} Inference cost (per 1M tokens; relative to the GPT-4o teacher at a 1:1 token ratio) and per-sample latency, measured over $50$ samples. \ours{} runs locally.}
    \label{table:efficiency}
    \setlength{\tabcolsep}{3pt}
    \resizebox{0.95\linewidth}{!}{%
    \begin{tabular}{l cc c cc}
        \toprule[1pt]
        \multirow{2.5}{*}{\textbf{Model}} & \multicolumn{2}{c}{\textbf{Cost / 1M}} & \multirow{2.5}{*}{\textbf{Rel.}} & \multicolumn{2}{c}{\textbf{Latency (s)}} \\
        \cmidrule(lr){2-3} \cmidrule(lr){5-6}
        & In & Out & & Anon. & Anon.+Inf. \\
        \midrule[0.75pt]
        Claude~Haiku~4.5 & \$1.00 & \$5.00 & 30.0\% & 0.81 & 5.64 \\
        Gemini~2.5~Flash & \$0.30 & \$2.50 & 14.0\% & \textbf{0.49} & \textbf{3.37} \\
        \midrule[0.25pt]
        \rowcolor{RowHighlight} \ours{} / SEAL & \$0.10 & \$0.10 & \textbf{1.0\%} & 0.64 & 13.76 \\
        \bottomrule[1pt]
    \end{tabular}
    }
\end{table}

\section{Per-iteration results}

\paragraph{Main and Hard splits.}
Tables~\ref{table:main-table-iter} and~\ref{table:hard-table-iter} report the
full self-refinement trajectories summarized in
Tables~\ref{table:main-table} and~\ref{table:hard-table}. Two
patterns hold across methods and splits. First, privacy improves
with refinement while utility declines monotonically, so the overall
score peaks at an intermediate round---for \ours{}, round~2 on both
splits ($0.396$ main, $0.294$ hard)---and the per-round trajectory
lets users pick an operating point along this trade-off. Second, the
methods differ in how far refinement carries them: the frontier
adversarial anonymizers improve privacy slowly and remain well above
$0.36$ (main) and $0.45$ (hard) after five rounds, whereas \ours{}
continues to gain, reaching $0.195$ on the main set (round~4) and
$0.332$ on the hard set (round~5). On the hard split \ours{}'s
privacy decreases at every round, and its per-attribute gains over
SEAL concentrate on the same attributes as on the main split,
consistent with the aggregate results in
Section~\ref{ss:main-results}.

\begin{table*}[h]
    \centering
    \small
\caption{\textbf{Per-iteration ablation of \ours{} components.}
Overall ($\uparrow$), privacy ($\downarrow$, attribute-inference
accuracy), and utility ($\uparrow$) at self-refinement rounds
R1--R3 (GLM-5 judge); R0 is the un-anonymized input and is omitted.
Highlighted: full model. Summarized at the best-Overall round in
Table~\ref{table:ablation}.}
    \label{table:ablation-iter}
    \resizebox{0.9\linewidth}{!}{%
    \setlength{\tabcolsep}{5pt}
    \begin{tabular}{l ccc ccc ccc}
        \toprule[1pt]
        \multirow{2}{*}{\textbf{Method}} & \multicolumn{3}{c}{Ovr.\,$\uparrow$} & \multicolumn{3}{c}{Priv.\,$\downarrow$} & \multicolumn{3}{c}{Util.\,$\uparrow$} \\
        \cmidrule(lr){2-4} \cmidrule(lr){5-7} \cmidrule(lr){8-10}
         & R1 & R2 & R3 & R1 & R2 & R3 & R1 & R2 & R3 \\
        \midrule[0.75pt]
        \multicolumn{10}{c}{\emph{Main dataset}} \\
        \midrule[0.25pt]
        SFT (no RL) & 0.172 & 0.216 & 0.279 & 0.483 & 0.436 & 0.388 & 0.951 & 0.919 & 0.905 \\
        \rowcolor{RowHighlight} \textbf{\ours{} (full)} & 0.282 & 0.351 & 0.353 & 0.378 & 0.281 & 0.234 & 0.892 & 0.804 & 0.730 \\
        \hspace{1em} w/o confidence reward & -0.038 & -0.017 & -0.045 & 0.639 & 0.626 & 0.643 & 0.993 & 0.992 & 0.991 \\
        \hspace{1em} w/o attacker feedback & 0.348 & 0.336 & 0.308 & 0.294 & 0.246 & 0.239 & 0.823 & 0.734 & 0.693 \\
        \hspace{1em} w/ truth-aligned reward & 0.276 & 0.324 & 0.346 & 0.383 & 0.301 & 0.252 & 0.893 & 0.809 & 0.753 \\
        \hspace{1em} w/o utility floor ($\tau{=}0$) & 0.376 & 0.360 & 0.303 & 0.272 & 0.212 & 0.215 & 0.815 & 0.702 & 0.650 \\
        \hspace{1em} fewer samples ($2$k vs $5$k) & 0.296 & 0.319 & 0.341 & 0.373 & 0.306 & 0.253 & 0.898 & 0.813 & 0.749 \\
        \hspace{1em} smaller group ($G{=}4$) & 0.212 & 0.299 & 0.333 & 0.453 & 0.373 & 0.333 & 0.943 & 0.900 & 0.870 \\
        \midrule[0.75pt]
        \multicolumn{10}{c}{\emph{Hard dataset}} \\
        \midrule[0.25pt]
        SFT (no RL) & 0.133 & 0.169 & 0.195 & 0.641 & 0.596 & 0.564 & 0.953 & 0.932 & 0.917 \\
        \rowcolor{RowHighlight} \textbf{\ours{} (full)} & 0.242 & 0.278 & 0.290 & 0.519 & 0.436 & 0.377 & 0.906 & 0.837 & 0.773 \\
        \hspace{1em} w/o confidence reward & -0.006 & 0.003 & 0.006 & 0.785 & 0.776 & 0.773 & 0.999 & 0.997 & 0.996 \\
        \hspace{1em} w/o attacker feedback & 0.249 & 0.265 & 0.262 & 0.478 & 0.387 & 0.356 & 0.861 & 0.761 & 0.717 \\
        \hspace{1em} w/ truth-aligned reward & 0.229 & 0.285 & 0.273 & 0.521 & 0.430 & 0.395 & 0.896 & 0.835 & 0.779 \\
        \hspace{1em} w/o utility floor ($\tau{=}0$) & 0.266 & 0.256 & 0.276 & 0.443 & 0.370 & 0.313 & 0.832 & 0.730 & 0.677 \\
        \hspace{1em} fewer samples ($2$k vs $5$k) & 0.253 & 0.306 & 0.300 & 0.510 & 0.426 & 0.382 & 0.907 & 0.851 & 0.789 \\
        \hspace{1em} smaller group ($G{=}4$) & 0.191 & 0.238 & 0.239 & 0.586 & 0.524 & 0.506 & 0.940 & 0.908 & 0.888 \\
        \bottomrule[1pt]
    \end{tabular}
    }
\end{table*}

\paragraph{Ablation.}
Table~\ref{table:ablation-iter} shows how each ablated variant
evolves over refinement, revealing dynamics that the best-round
summary in Table~\ref{table:ablation} compresses. Without the
confidence reward, refinement makes no progress at all: privacy
stays at $0.626$--$0.643$ (main) and $0.773$--$0.785$ (hard) across
all three rounds, and the overall score never rises above $0.006$.
The utility floor shapes the trajectory rather than a single round:
with $\tau{=}0$ the policy reaches stronger raw privacy at every
matched round, but utility falls steeply with each refinement
($0.815{\to}0.650$ main, $0.832{\to}0.677$ hard) and the overall
score declines from R1 to R3 ($0.376{\to}0.303$ on main), whereas
the full model's overall score rises ($0.282{\to}0.353$)---the floor
trades some immediate privacy for refinement that remains
worthwhile. The remaining variants---no attacker feedback, fewer
training samples, and the smaller group $G{=}4$---track the full
model's shape but sit at consistently weaker privacy, with $G{=}4$
the furthest behind at every round.

\paragraph{Judge robustness.}
Table~\ref{table:judge-robustness-iter} extends the best-round
comparison of Table~\ref{table:judge-robustness} to every
self-refinement step. \ours{} leads SEAL on the overall score in all
but three of the $36$ judge$\times$split$\times$round cells: the
first round on the hard split under GLM-5 ($0.239$ vs.\ $0.248$) and
Qwen3 ($0.244$ vs.\ $0.253$), before self-refinement takes effect,
and the final round on the main split under GPT-OSS, where the two
methods are within $0.003$. Absolute scores shift across judges,
mainly through stricter utility assessments, but the per-round
ordering is preserved, indicating that the advantage is neither an
artifact of the evaluator nor of selecting a favorable round.

\section{Qualitative comparison}
\label{qualitative}

Table~\ref{tab:qualitative} traces how each method rewrites a
single comment, ``rent eats up most of my paycheck these days'',
whose author has low income. Azure finds no explicit identifier and
leaves the text unchanged. Dipper paraphrases without an
anonymization objective and distorts the meaning (``My rent goes a
long way these days''), consistent with its large utility loss in
Tables~\ref{table:main-table-iter} and~\ref{table:hard-table-iter}. The frontier
adversarial anonymizers soften ``paycheck'' into ``income'' or
``expenses'' but still convey that the cost of living strains the
author, leaving the low-income cue largely intact. SEAL
over-generalizes across rounds, drifting into stilted,
near-meaningless phrasing (``a significant amount of funds is being
allocated to a particular area''). \ours{} instead removes the
personal-income cue while keeping a natural, coherent sentence
(``It can be expensive to pay for a place to live''). This mirrors
the quantitative picture: \ours{} reaches the strongest privacy
while preserving readable, faithful text.

\begin{table*}[t]
    \centering
    \footnotesize
    \caption{\textbf{Per-iteration anonymization results on the main
dataset.} Overall ($\uparrow$), privacy ($\downarrow$,
attribute-inference accuracy), and utility ($\uparrow$) per
self-refinement round, judged by GLM-5, with per-attribute inference
accuracy. Overall is the relative privacy improvement minus the
relative utility loss, normalized by the original privacy
($P_{\text{orig}}{=}0.620$). Top panel: the five-round trajectories
of the three adversarial-anonymization baselines; bottom panel: the
original text, Azure, Dipper, and the trajectories of SEAL and
\ours{}. \textbf{Bold}/\underline{underline}: best/second-best per row. \ours{} attains
both the best overall score ($0.396$, round~2) and the lowest
inference accuracy ($0.195$, round~4). Summarized in
Table~\ref{table:main-table} and plotted in
Figure~\ref{fig:tradeoff}.}
    \label{table:main-table-iter}
    \vspace{0.05in}
    \setlength{\tabcolsep}{1.5pt}
    \resizebox{\textwidth}{!}{%
\begin{tabular}{l ccccc ccccc ccccc}
\toprule[1pt]
\multirow{2.5}{*}{\textbf{Metric}} & \multicolumn{5}{c}{\textbf{Adv.\ Anon.\ Gemini}} & \multicolumn{5}{c}{\textbf{Adv.\ Anon.\ Claude}} & \multicolumn{5}{c}{\textbf{Adv.\ Anon.\ Qwen3}} \\
\cmidrule(lr){2-6} \cmidrule(lr){7-11} \cmidrule(lr){12-16}
 & \scriptsize{iter 1} & \scriptsize{iter 2} & \scriptsize{iter 3} & \scriptsize{iter 4} & \scriptsize{iter 5} & \scriptsize{iter 1} & \scriptsize{iter 2} & \scriptsize{iter 3} & \scriptsize{iter 4} & \scriptsize{iter 5} & \scriptsize{iter 1} & \scriptsize{iter 2} & \scriptsize{iter 3} & \scriptsize{iter 4} & \scriptsize{iter 5} \\
\midrule[0.75pt]
\rowcolor{RowHighlight} \textbf{Overall $\uparrow$} & 0.169 & 0.202 & 0.259 & 0.262 & 0.284 & 0.148 & 0.202 & 0.246 & 0.239 & 0.292 & 0.089 & 0.188 & 0.165 & 0.207 & 0.175 \\
\midrule[0.50pt]
\rowcolor{RowHighlight} \textbf{Privacy $\downarrow$} & 0.473 & 0.443 & 0.397 & 0.383 & 0.367 & 0.488 & 0.438 & 0.400 & 0.400 & 0.365 & 0.544 & 0.470 & 0.480 & 0.454 & 0.467 \\
\midrule[0.25pt]
\hspace{1em} Age & 0.700 & 0.703 & 0.693 & 0.703 & 0.720 & 0.743 & 0.713 & 0.683 & 0.713 & 0.614 & 0.733 & 0.733 & 0.733 & 0.750 & 0.752 \\
\hspace{1em} Edu & 0.640 & 0.635 & 0.576 & 0.550 & 0.521 & 0.630 & 0.621 & 0.514 & 0.531 & 0.510 & 0.654 & 0.602 & 0.616 & 0.576 & 0.619 \\
\hspace{1em} Gnd & 0.639 & 0.623 & 0.525 & 0.525 & 0.508 & 0.672 & 0.541 & 0.607 & 0.656 & 0.590 & 0.639 & 0.607 & 0.672 & 0.623 & 0.689 \\
\hspace{1em} Inc & 0.541 & 0.592 & 0.571 & 0.561 & 0.510 & 0.571 & 0.561 & 0.510 & 0.541 & 0.490 & 0.653 & 0.561 & 0.571 & 0.561 & 0.592 \\
\hspace{1em} Loc & 0.127 & 0.082 & 0.060 & 0.067 & 0.067 & 0.179 & 0.142 & 0.127 & 0.114 & 0.090 & 0.209 & 0.134 & 0.127 & 0.097 & 0.112 \\
\hspace{1em} Mar & 0.662 & 0.627 & 0.560 & 0.587 & 0.554 & 0.653 & 0.627 & 0.613 & 0.640 & 0.608 & 0.747 & 0.667 & 0.653 & 0.693 & 0.680 \\
\hspace{1em} Occ & 0.320 & 0.246 & 0.197 & 0.173 & 0.163 & 0.333 & 0.249 & 0.223 & 0.185 & 0.167 & 0.429 & 0.325 & 0.334 & 0.289 & 0.280 \\
\hspace{1em} PoB & 0.179 & 0.214 & 0.214 & 0.107 & 0.143 & 0.143 & 0.179 & 0.179 & 0.214 & 0.179 & 0.143 & 0.071 & 0.071 & 0.143 & 0.143 \\
\midrule[0.50pt]
\rowcolor{RowHighlight} \textbf{Utility $\uparrow$} & 0.932 & 0.916 & 0.899 & 0.880 & 0.876 & 0.936 & 0.908 & 0.891 & 0.884 & 0.880 & \textbf{0.967} & 0.946 & 0.940 & 0.940 & 0.928 \\
\midrule[0.25pt]
\hspace{1em} Mean & 0.893 & 0.849 & 0.804 & 0.774 & 0.750 & 0.889 & 0.824 & 0.793 & 0.768 & 0.755 & \underline{0.930} & 0.898 & 0.879 & 0.873 & 0.861 \\
\hspace{1em} Read & 0.998 & 0.995 & 0.993 & 0.989 & 0.988 & 0.998 & 0.995 & 0.993 & 0.994 & 0.991 & \textbf{0.999} & \underline{0.998} & 0.997 & 0.998 & 0.998 \\
\hspace{1em} Hall & 0.906 & 0.905 & 0.900 & 0.876 & 0.888 & 0.920 & 0.906 & 0.888 & 0.888 & 0.893 & \underline{0.971} & 0.942 & 0.943 & 0.947 & 0.924 \\
\bottomrule[1pt]
\end{tabular}
    }
    \vspace{2pt}
    \resizebox{\textwidth}{!}{%
\begin{tabular}{l c cc ccccc ccccc}
\toprule[1pt]
\multirow{2.5}{*}{\textbf{Metric}} & \multirow{2.5}{*}{\textbf{Original}} & \multirow{2.5}{*}{\textbf{Azure}} & \multirow{2.5}{*}{\textbf{Dipper}} & \multicolumn{5}{c}{\textbf{SEAL}} & \multicolumn{5}{c}{\textbf{\ours{}} (Ours)} \\
\cmidrule(lr){5-9} \cmidrule(lr){10-14}
 & & & & \scriptsize{iter 1} & \scriptsize{iter 2} & \scriptsize{iter 3} & \scriptsize{iter 4} & \scriptsize{iter 5} & \scriptsize{iter 1} & \scriptsize{iter 2} & \scriptsize{iter 3} & \scriptsize{iter 4} & \scriptsize{iter 5} \\
\midrule[0.75pt]
\rowcolor{RowHighlight} \textbf{Overall $\uparrow$} & - & -0.033 & -0.018 & 0.320 & 0.368 & 0.297 & 0.304 & 0.276 & \underline{0.382} & \textbf{0.396} & 0.315 & 0.368 & 0.342 \\
\midrule[0.50pt]
\rowcolor{RowHighlight} \textbf{Privacy $\downarrow$} & 0.620 & 0.610 & 0.538 & 0.348 & 0.262 & 0.249 & 0.244 & 0.249 & 0.319 & 0.260 & 0.252 & \textbf{0.195} & \underline{0.198} \\
\midrule[0.25pt]
\hspace{1em} Age & 0.762 & 0.802 & 0.614 & 0.624 & 0.551 & 0.567 & \underline{0.500} & 0.538 & 0.550 & 0.600 & 0.650 & \textbf{0.300} & 0.500 \\
\hspace{1em} Edu & 0.645 & 0.667 & 0.718 & 0.502 & 0.416 & 0.313 & 0.325 & 0.345 & 0.391 & 0.266 & 0.297 & \textbf{0.203} & \underline{0.203} \\
\hspace{1em} Gnd & 0.883 & 0.820 & 0.721 & 0.607 & 0.450 & 0.345 & 0.417 & 0.386 & 0.583 & 0.417 & 0.333 & \textbf{0.250} & \underline{0.250} \\
\hspace{1em} Inc & 0.561 & 0.598 & 0.531 & 0.485 & \textbf{0.457} & 0.559 & 0.521 & 0.516 & 0.571 & 0.607 & 0.536 & 0.536 & \underline{0.464} \\
\hspace{1em} Loc & 0.470 & 0.362 & 0.300 & 0.030 & \textbf{0.015} & \underline{0.023} & 0.023 & 0.033 & 0.100 & 0.050 & 0.050 & 0.050 & 0.050 \\
\hspace{1em} Mar & 0.635 & 0.676 & 0.541 & 0.600 & 0.324 & 0.479 & 0.425 & 0.384 & 0.273 & 0.409 & 0.318 & \textbf{0.182} & \underline{0.182} \\
\hspace{1em} Occ & 0.578 & 0.542 & 0.448 & 0.139 & 0.069 & \textbf{0.045} & \underline{0.048} & 0.055 & 0.168 & 0.069 & 0.082 & 0.083 & 0.073 \\
\hspace{1em} PoB & 0.321 & 0.286 & 0.167 & \textbf{0.071} & \underline{0.071} & 0.107 & 0.125 & 0.107 & 0.375 & 0.500 & 0.125 & 0.500 & 0.500 \\
\midrule[0.50pt]
\rowcolor{RowHighlight} \textbf{Utility $\uparrow$} & 1.000 & \underline{0.951} & 0.850 & 0.882 & 0.791 & 0.699 & 0.698 & 0.679 & 0.895 & 0.815 & 0.721 & 0.683 & 0.662 \\
\midrule[0.25pt]
\hspace{1em} Mean & 1.000 & \textbf{0.936} & 0.848 & 0.771 & 0.598 & 0.470 & 0.474 & 0.453 & 0.765 & 0.609 & 0.496 & 0.421 & 0.370 \\
\hspace{1em} Read & 1.000 & 0.922 & 0.950 & 0.997 & 0.991 & 0.984 & 0.983 & 0.954 & 0.995 & 0.993 & 0.985 & 0.982 & 0.974 \\
\hspace{1em} Hall & 1.000 & \textbf{0.994} & 0.752 & 0.876 & 0.783 & 0.643 & 0.636 & 0.630 & 0.927 & 0.844 & 0.682 & 0.646 & 0.641 \\
\bottomrule[1pt]
\end{tabular}
    }
\end{table*}

\begin{table*}[t]
    \centering
    \footnotesize
    \caption{\textbf{Per-iteration anonymization results on the hard
dataset.} Overall ($\uparrow$), privacy ($\downarrow$,
attribute-inference accuracy), and utility ($\uparrow$) per
self-refinement round, judged by GLM-5, with per-attribute inference
accuracy. Overall is the relative privacy improvement minus the
relative utility loss, normalized by the original privacy
($P_{\text{orig}}{=}0.781$). Top panel: the five-round trajectories
of the three adversarial-anonymization baselines; bottom panel: the
original text, Azure, Dipper, and the trajectories of SEAL and
\ours{}. \textbf{Bold}/\underline{underline}: best/second-best per row. \ours{}'s
inference accuracy decreases at every round, reaching $0.332$, the
lowest of any method. Summarized in Table~\ref{table:hard-table} and
plotted in Figure~\ref{fig:tradeoff}.}
    \label{table:hard-table-iter}
    \vspace{0.05in}
    \setlength{\tabcolsep}{1.5pt}
    \resizebox{\textwidth}{!}{%
\begin{tabular}{l ccccc ccccc ccccc}
\toprule[1pt]
\multirow{2.5}{*}{\textbf{Metric}} & \multicolumn{5}{c}{\textbf{Adv.\ Anon.\ Gemini}} & \multicolumn{5}{c}{\textbf{Adv.\ Anon.\ Claude}} & \multicolumn{5}{c}{\textbf{Adv.\ Anon.\ Qwen3}} \\
\cmidrule(lr){2-6} \cmidrule(lr){7-11} \cmidrule(lr){12-16}
 & \scriptsize{iter 1} & \scriptsize{iter 2} & \scriptsize{iter 3} & \scriptsize{iter 4} & \scriptsize{iter 5} & \scriptsize{iter 1} & \scriptsize{iter 2} & \scriptsize{iter 3} & \scriptsize{iter 4} & \scriptsize{iter 5} & \scriptsize{iter 1} & \scriptsize{iter 2} & \scriptsize{iter 3} & \scriptsize{iter 4} & \scriptsize{iter 5} \\
\midrule[0.75pt]
\rowcolor{RowHighlight} \textbf{Overall $\uparrow$} & 0.194 & 0.229 & 0.276 & 0.291 & \textbf{0.301} & 0.188 & 0.253 & 0.253 & 0.278 & \underline{0.299} & 0.181 & 0.238 & 0.258 & 0.265 & 0.262 \\
\midrule[0.50pt]
\rowcolor{RowHighlight} \textbf{Privacy $\downarrow$} & 0.593 & 0.550 & 0.499 & 0.478 & 0.465 & 0.595 & 0.526 & 0.508 & 0.473 & 0.454 & 0.608 & 0.553 & 0.526 & 0.517 & 0.517 \\
\midrule[0.25pt]
\hspace{1em} Age & 0.886 & 0.847 & 0.780 & 0.750 & 0.738 & 0.868 & 0.813 & 0.796 & 0.751 & 0.717 & 0.899 & 0.836 & 0.818 & 0.800 & 0.811 \\
\hspace{1em} Edu & 0.825 & 0.785 & 0.736 & 0.704 & 0.665 & 0.827 & 0.757 & 0.719 & 0.676 & 0.674 & 0.821 & 0.788 & 0.765 & 0.744 & 0.755 \\
\hspace{1em} Gnd & 0.690 & 0.762 & 0.714 & 0.643 & 0.738 & 0.738 & 0.643 & 0.585 & 0.643 & 0.571 & 0.738 & 0.619 & 0.643 & 0.595 & 0.610 \\
\hspace{1em} Inc & 0.540 & 0.607 & 0.546 & 0.503 & 0.520 & 0.543 & 0.543 & 0.564 & 0.545 & 0.529 & 0.547 & 0.562 & 0.538 & 0.573 & 0.566 \\
\hspace{1em} Loc & 0.280 & 0.230 & 0.221 & 0.219 & 0.219 & 0.297 & 0.245 & 0.228 & 0.219 & 0.217 & 0.317 & 0.247 & 0.229 & 0.227 & 0.235 \\
\hspace{1em} Mar & 0.800 & 0.770 & 0.747 & 0.747 & 0.662 & 0.800 & 0.733 & 0.787 & 0.716 & 0.680 & 0.813 & 0.773 & 0.760 & 0.733 & 0.743 \\
\hspace{1em} Occ & 0.541 & 0.413 & 0.323 & 0.318 & 0.291 & 0.534 & 0.387 & 0.368 & 0.300 & 0.265 & 0.575 & 0.489 & 0.421 & 0.388 & 0.357 \\
\hspace{1em} PoB & 0.299 & 0.190 & 0.179 & 0.162 & 0.155 & 0.321 & 0.218 & 0.166 & 0.140 & \textbf{0.118} & 0.300 & 0.187 & 0.160 & 0.168 & 0.170 \\
\midrule[0.50pt]
\rowcolor{RowHighlight} \textbf{Utility $\uparrow$} & \underline{0.953} & 0.932 & 0.915 & 0.903 & 0.896 & 0.950 & 0.926 & 0.904 & 0.884 & 0.879 & \textbf{0.960} & 0.946 & 0.931 & 0.927 & 0.924 \\
\midrule[0.25pt]
\hspace{1em} Mean & 0.879 & 0.821 & 0.775 & 0.754 & 0.722 & 0.873 & 0.799 & 0.753 & 0.699 & 0.678 & \textbf{0.906} & 0.861 & 0.830 & 0.811 & 0.794 \\
\hspace{1em} Read & 0.999 & 0.998 & 0.999 & 0.998 & 0.995 & \textbf{1.000} & 0.999 & 0.998 & 0.998 & 0.995 & \underline{1.000} & 0.999 & 0.999 & 0.999 & 0.999 \\
\hspace{1em} Hall & \underline{0.980} & 0.978 & 0.972 & 0.956 & 0.972 & 0.978 & 0.980 & 0.962 & 0.954 & 0.966 & 0.974 & 0.978 & 0.964 & 0.972 & 0.978 \\
\bottomrule[1pt]
\end{tabular}
    }
    \vspace{2pt}
    \resizebox{\textwidth}{!}{%
\begin{tabular}{l c cc ccccc ccccc}
\toprule[1pt]
\multirow{2.5}{*}{\textbf{Metric}} & \multirow{2.5}{*}{\textbf{Original}} & \multirow{2.5}{*}{\textbf{Azure}} & \multirow{2.5}{*}{\textbf{Dipper}} & \multicolumn{5}{c}{\textbf{SEAL} } & \multicolumn{5}{c}{\textbf{\ours{}} (Ours)} \\
\cmidrule(lr){5-9} \cmidrule(lr){10-14}
 & & & & \scriptsize{iter 1} & \scriptsize{iter 2} & \scriptsize{iter 3} & \scriptsize{iter 4} & \scriptsize{iter 5} & \scriptsize{iter 1} & \scriptsize{iter 2} & \scriptsize{iter 3} & \scriptsize{iter 4} & \scriptsize{iter 5} \\
\midrule[0.75pt]
\rowcolor{RowHighlight} \textbf{Overall $\uparrow$} & - & -0.023 & -0.035 & 0.248 & 0.266 & 0.254 & 0.249 & 0.242 & 0.239 & 0.294 & 0.294 & 0.279 & 0.281 \\
\midrule[0.50pt]
\rowcolor{RowHighlight} \textbf{Privacy $\downarrow$} & 0.781 & 0.734 & 0.707 & 0.497 & 0.420 & 0.369 & 0.385 & 0.377 & 0.521 & 0.430 & 0.379 & \underline{0.356} & \textbf{0.332} \\
\midrule[0.25pt]
\hspace{1em} Age & 0.912 & 0.868 & 0.789 & 0.744 & 0.655 & 0.589 & 0.623 & 0.606 & 0.804 & 0.679 & 0.608 & \underline{0.580} & \textbf{0.534} \\
\hspace{1em} Edu & 0.834 & 0.802 & 0.828 & 0.714 & 0.616 & 0.512 & 0.527 & 0.550 & 0.723 & 0.593 & 0.501 & \underline{0.472} & \textbf{0.417} \\
\hspace{1em} Gnd & 1.000 & 0.838 & 0.952 & 0.738 & 0.667 & 0.619 & 0.571 & 0.595 & 0.810 & 0.595 & 0.571 & \textbf{0.500} & \underline{0.500} \\
\hspace{1em} Inc & 0.520 & 0.541 & 0.484 & 0.538 & 0.508 & \underline{0.467} & 0.506 & 0.470 & 0.552 & 0.540 & 0.487 & 0.475 & \textbf{0.457} \\
\hspace{1em} Loc & 0.891 & 0.796 & 0.761 & 0.240 & 0.203 & 0.203 & \textbf{0.200} & \underline{0.201} & 0.273 & 0.221 & 0.212 & 0.208 & 0.209 \\
\hspace{1em} Mar & 0.907 & 0.826 & 0.840 & 0.693 & 0.667 & 0.613 & 0.627 & 0.589 & 0.760 & 0.613 & 0.627 & \underline{0.560} & \textbf{0.527} \\
\hspace{1em} Occ & 0.624 & 0.566 & 0.555 & 0.360 & 0.214 & 0.147 & 0.155 & 0.137 & 0.339 & 0.217 & 0.169 & \underline{0.124} & \textbf{0.117} \\
\hspace{1em} PoB & 0.908 & 0.831 & 0.806 & 0.184 & 0.126 & 0.124 & 0.123 & 0.125 & 0.241 & 0.165 & 0.126 & 0.126 & \underline{0.120} \\
\midrule[0.50pt]
\rowcolor{RowHighlight} \textbf{Utility $\uparrow$} & 1.000 & 0.916 & 0.870 & 0.884 & 0.804 & 0.727 & 0.741 & 0.724 & 0.906 & 0.844 & 0.780 & 0.735 & 0.706 \\
\midrule[0.25pt]
\hspace{1em} Mean & 1.000 & \underline{0.898} & 0.893 & 0.727 & 0.572 & 0.449 & 0.457 & 0.435 & 0.755 & 0.600 & 0.483 & 0.414 & 0.360 \\
\hspace{1em} Read & 1.000 & 0.850 & 0.970 & 0.998 & 0.993 & 0.981 & 0.980 & 0.969 & 0.996 & 0.985 & 0.966 & 0.954 & 0.947 \\
\hspace{1em} Hall & 1.000 & \textbf{1.000} & 0.748 & 0.928 & 0.848 & 0.750 & 0.786 & 0.768 & 0.968 & 0.948 & 0.890 & 0.836 & 0.811 \\
\bottomrule[1pt]
\end{tabular}
    }
\end{table*}

\begin{table*}[p]
    \centering
    \small
    \caption{\textbf{Per-iteration judge robustness.} Overall
($\uparrow$), privacy ($\downarrow$, attribute-inference accuracy),
and utility ($\uparrow$) at each of six self-refinement rounds for
SEAL and \ours{}, scored by three independent judges on both splits.
Bold: better of the two methods per iteration and metric. \ours{}
leads on Overall at nearly every round under every judge. Summarized
at the best-Overall round in Table~\ref{table:judge-robustness}.}
    \label{table:judge-robustness-iter}
    \setlength{\tabcolsep}{7pt}
    \resizebox{0.9\linewidth}{!}{%
    \begin{tabular}{lll cccccc}
        \toprule[1pt]
        \textbf{Metric} & \textbf{Judge} & \textbf{Method} & it1 & it2 & it3 & it4 & it5 & it6 \\
        \midrule[0.75pt]
        \multicolumn{9}{c}{\emph{Main dataset}} \\
        \midrule[0.25pt]
        \multirow{6}{*}{Overall $\uparrow$} & \multirow{2}{*}{GLM-5} & SEAL & 0.320 & 0.368 & 0.297 & 0.304 & 0.276 & 0.227 \\
         & & \ours{} & \textbf{0.382} & \textbf{0.396} & \textbf{0.315} & \textbf{0.368} & \textbf{0.342} & \textbf{0.293} \\
         & \multirow{2}{*}{Qwen3} & SEAL & 0.249 & 0.320 & 0.342 & 0.321 & 0.300 & 0.268 \\
         & & \ours{} & \textbf{0.319} & \textbf{0.400} & \textbf{0.390} & \textbf{0.366} & \textbf{0.384} & \textbf{0.376} \\
         & \multirow{2}{*}{GPT-OSS} & SEAL & 0.291 & 0.257 & 0.222 & 0.221 & 0.200 & \textbf{0.202} \\
         & & \ours{} & \textbf{0.319} & \textbf{0.312} & \textbf{0.237} & \textbf{0.245} & \textbf{0.208} & 0.199 \\
        \addlinespace[3pt]
        \multirow{6}{*}{Privacy $\downarrow$} & \multirow{2}{*}{GLM-5} & SEAL & 0.348 & 0.262 & \textbf{0.249} & 0.244 & 0.249 & 0.268 \\
         & & \ours{} & \textbf{0.319} & \textbf{0.260} & 0.252 & \textbf{0.195} & \textbf{0.198} & \textbf{0.207} \\
         & \multirow{2}{*}{Qwen3} & SEAL & 0.375 & 0.286 & 0.236 & 0.246 & 0.245 & 0.255 \\
         & & \ours{} & \textbf{0.347} & \textbf{0.258} & \textbf{0.234} & \textbf{0.227} & \textbf{0.202} & \textbf{0.194} \\
         & \multirow{2}{*}{GPT-OSS} & SEAL & 0.317 & 0.254 & \textbf{0.214} & 0.217 & 0.222 & 0.211 \\
         & & \ours{} & \textbf{0.309} & \textbf{0.243} & 0.232 & \textbf{0.198} & \textbf{0.195} & \textbf{0.181} \\
        \addlinespace[3pt]
        \multirow{6}{*}{Utility $\uparrow$} & \multirow{2}{*}{GLM-5} & SEAL & 0.882 & 0.791 & 0.699 & \textbf{0.698} & \textbf{0.679} & \textbf{0.660} \\
         & & \ours{} & \textbf{0.895} & \textbf{0.815} & \textbf{0.721} & 0.683 & 0.662 & 0.627 \\
         & \multirow{2}{*}{Qwen3} & SEAL & 0.853 & 0.782 & 0.723 & 0.718 & 0.694 & 0.679 \\
         & & \ours{} & \textbf{0.879} & \textbf{0.817} & \textbf{0.767} & \textbf{0.733} & \textbf{0.710} & \textbf{0.689} \\
         & \multirow{2}{*}{GPT-OSS} & SEAL & 0.802 & 0.667 & 0.567 & \textbf{0.570} & \textbf{0.558} & \textbf{0.542} \\
         & & \ours{} & \textbf{0.817} & \textbf{0.704} & \textbf{0.611} & 0.564 & 0.523 & 0.491 \\
        \midrule[0.75pt]
        \multicolumn{9}{c}{\emph{Hard dataset}} \\
        \midrule[0.25pt]
        \multirow{6}{*}{Overall $\uparrow$} & \multirow{2}{*}{GLM-5} & SEAL & \textbf{0.248} & 0.266 & 0.254 & 0.249 & 0.242 & 0.232 \\
         & & \ours{} & 0.239 & \textbf{0.294} & \textbf{0.294} & \textbf{0.279} & \textbf{0.281} & \textbf{0.264} \\
         & \multirow{2}{*}{Qwen3} & SEAL & \textbf{0.253} & 0.300 & 0.285 & 0.297 & 0.273 & 0.264 \\
         & & \ours{} & 0.244 & \textbf{0.327} & \textbf{0.336} & \textbf{0.335} & \textbf{0.346} & \textbf{0.329} \\
         & \multirow{2}{*}{GPT-OSS} & SEAL & 0.286 & 0.296 & 0.299 & 0.281 & 0.272 & 0.266 \\
         & & \ours{} & \textbf{0.306} & \textbf{0.357} & \textbf{0.352} & \textbf{0.308} & \textbf{0.298} & \textbf{0.274} \\
        \addlinespace[3pt]
        \multirow{6}{*}{Privacy $\downarrow$} & \multirow{2}{*}{GLM-5} & SEAL & \textbf{0.497} & \textbf{0.420} & \textbf{0.369} & 0.385 & 0.377 & 0.376 \\
         & & \ours{} & 0.521 & 0.430 & 0.379 & \textbf{0.356} & \textbf{0.332} & \textbf{0.326} \\
         & \multirow{2}{*}{Qwen3} & SEAL & \textbf{0.472} & \textbf{0.378} & \textbf{0.340} & 0.329 & 0.340 & 0.338 \\
         & & \ours{} & 0.497 & 0.388 & 0.340 & \textbf{0.318} & \textbf{0.293} & \textbf{0.283} \\
         & \multirow{2}{*}{GPT-OSS} & SEAL & \textbf{0.417} & 0.326 & \textbf{0.254} & 0.268 & 0.269 & 0.266 \\
         & & \ours{} & 0.425 & \textbf{0.318} & 0.270 & \textbf{0.254} & \textbf{0.230} & \textbf{0.216} \\
        \addlinespace[3pt]
        \multirow{6}{*}{Utility $\uparrow$} & \multirow{2}{*}{GLM-5} & SEAL & 0.884 & 0.804 & 0.727 & \textbf{0.741} & \textbf{0.724} & \textbf{0.714} \\
         & & \ours{} & \textbf{0.906} & \textbf{0.844} & \textbf{0.780} & 0.735 & 0.706 & 0.682 \\
         & \multirow{2}{*}{Qwen3} & SEAL & 0.858 & 0.784 & 0.720 & 0.718 & 0.708 & \textbf{0.696} \\
         & & \ours{} & \textbf{0.881} & \textbf{0.824} & \textbf{0.772} & \textbf{0.742} & \textbf{0.721} & 0.691 \\
         & \multirow{2}{*}{GPT-OSS} & SEAL & 0.820 & 0.713 & 0.624 & 0.624 & \textbf{0.617} & \textbf{0.607} \\
         & & \ours{} & \textbf{0.850} & \textbf{0.765} & \textbf{0.697} & \textbf{0.633} & 0.593 & 0.550 \\
        \bottomrule[1pt]
    \end{tabular}
    }
\end{table*}

\begin{table*}[p]
    \centering
    \footnotesize
    \caption{\textbf{Round-by-round anonymization of the same comment by each method}, where the goal is to hide that the author has low income (their rent consumes most of their paycheck). Azure leaves the text unchanged and Dipper paraphrases but distorts the meaning; the frontier adversarial anonymizers soften the wording but keep the financial-strain signal; SEAL over-generalizes into stilted, near-meaningless phrasing, whereas \ours{} removes the income cue while keeping a natural, coherent sentence. Identical consecutive rounds are merged.}
    \label{tab:qualitative}
    \vspace{0.05in}
    \begin{tabularx}{\textwidth}{l|c|X}
        \toprule[1pt]
        \textbf{Method} & \textbf{Round} & \textbf{Text} \\
        \midrule[0.75pt]
        Original & -- & rent eats up most of my paycheck these days \\
        \midrule[0.50pt]
        Azure & 1--5 & rent eats up most of my paycheck these days \\
        \midrule[0.25pt]
        Dipper & 1 & My rent goes a long way these days \\
        \midrule[0.25pt]
        \multirow{4}{*}{AA (Gemini)} & 1 & Rent eats up most of my income these days \\
                                     & 2 & Housing costs eat up most of my income these days \\
                                     & 3--4 & Housing costs are a significant expense these days \\
                                     & 5 & The cost of living is a significant expense these days \\
        \midrule[0.25pt]
        \multirow{3}{*}{AA (Claude)} & 1 & rent eats up a significant portion of my paycheck these days \\
                                     & 2 & housing costs take up a significant portion of my paycheck these days \\
                                     & 3--5 & housing costs take up a significant portion of my expenses these days \\
        \midrule[0.25pt]
        AA (Qwen3) & 1--5 & Housing costs eat up most of my income these days \\
        \midrule[0.25pt]
        \multirow{4}{*}{SEAL} & 1 & rent takes up a lot of money right now \\
                              & 2 & a lot of money is being spent on something right now \\
                              & 3--4 & a lot of money is being spent on something \\
                              & 5 & a significant amount of funds is being allocated to a particular area \\
        \midrule[0.25pt]
        \multirow{5}{*}{\ours{}} & 1 & a lot of money is going towards housing \\
                                 & 2 & a lot of money is going towards housing costs \\
                                 & 3 & Housing costs are a significant expense. \\
                                 & 4 & It can be expensive to pay for a place to live. \\
                                 & 5 & It can be expensive to find a place to live. \\
        \bottomrule[1pt]
    \end{tabularx}
\end{table*}

\end{document}